\documentclass[sigconf, review=False, nonacm]{acmart}

\usepackage{multirow, dsfont, enumitem}

\AtBeginDocument{%
  \providecommand\BibTeX{{%
    \normalfont B\kern-0.5em{\scshape i\kern-0.25em b}\kern-0.8em\TeX}}}

\usepackage{subfig}
\usepackage{graphicx}
\usepackage[percent]{overpic}
\usepackage[countmax]{subfloat}
\usepackage{stfloats}
\usepackage{mathtools}
\usepackage{hyperref}
\usepackage{amsmath}

\usepackage{amsmath}
\usepackage{booktabs}
\usepackage{amsmath}
\usepackage{tabularx}

\graphicspath{ {./images/} }

\begin{document}

\def\be{\textbf{e}}

\def\bE{\textbf{E}}
\def\bF{\textbf{F}}
\def\bQ{\textbf{Q}}
\def\bK{\textbf{K}}
\def\bV{\textbf{V}}
\def\bS{\textbf{S}}

\def\bW{\textbf{W}}
\def\bv{\textbf{v}}
\def\bv{\textbf{EP}}

\def\cV{\mathcal{V}}

\title{adSformers: Personalization from Short-Term Sequences and Diversity of Representations in Etsy Ads}

\author{Alaa Awad, Denisa Roberts}
\authornote{Denisa Roberts is the corresponding author.}
\author{Eden Dolev, Andrea Heyman, Zahra Ebrahimzadeh, Zoe Weil, Marcin Mejran, Vaibhav Malpani, Mahir Yavuz}
\affiliation{%
  \institution{Etsy}
  \streetaddress{117 Adams St}
  \city{Brooklyn}
  \state{NY}
  \country{USA}
  \postcode{11201}
}
\email{{aawad, denisaroberts, edolev, aheyman}@etsy.com}

\renewcommand{\shortauthors}{Awad, Roberts, Dolev, and Heyman, et al.}

\begin{abstract}

In this article, we present a general approach to personalizing ads through encoding and learning from variable-length sequences of recent user actions and diverse representations. To this end we introduce a three-component module called the adSformer diversifiable personalization module (ADPM) that learns a dynamic user representation. We illustrate the module's effectiveness and flexibility by personalizing the Click-Through Rate (CTR) and Post-Click Conversion Rate (PCCVR) models used in sponsored search. The first component of the ADPM, the adSformer encoder, includes a novel adSformer block which learns the most salient sequence signals. ADPM's second component enriches the learned signal through visual, multimodal, and other pretrained representations. Lastly, the third ADPM "learned on the fly" component further diversifies the signal encoded in the dynamic user representation. The ADPM-personalized CTR and PCCVR models, henceforth referred to as adSformer CTR and adSformer PCCVR, outperform the CTR and PCCVR production baselines by $+2.66\%$ and $+2.42\%$, respectively, in offline Area Under the Receiver Operating Characteristic Curve (ROC-AUC). Following the robust online gains in A/B tests, Etsy Ads deployed the ADPM-personalized sponsored search system to $100\%$ of traffic as of February 2023.

\end{abstract}


\keywords{Personalization, Sponsored Search, Online Advertising, Neural Networks, Transformers, Representation Learning, Multimodality, Ranking, Bidding, Ads Computer Vision}

\maketitle

\begin{figure}[htp]
    \centering
    \subfloat[\centering non-personalized ranking]{{\includegraphics[width=\linewidth]{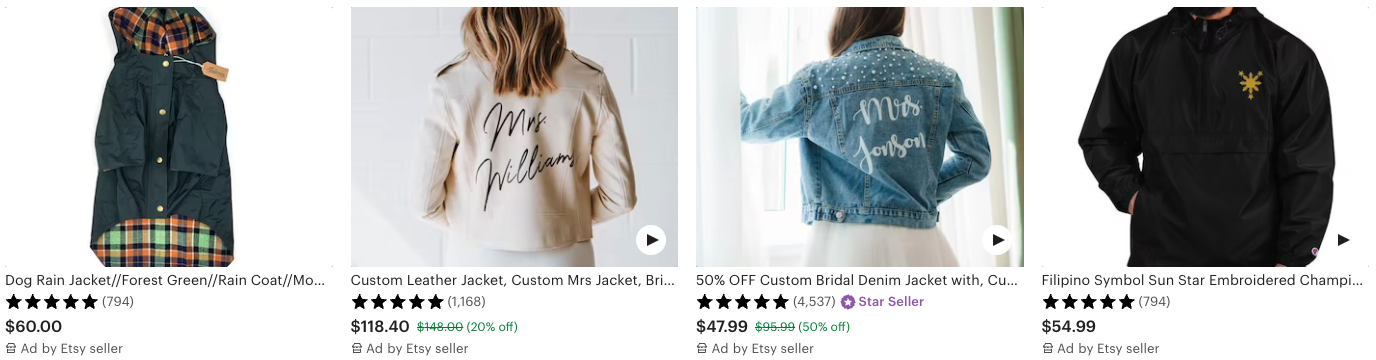} }}%
     \hspace{1.5cm}
    \subfloat[\centering with ADPM-personalized ranking]{{\includegraphics[width=\linewidth]{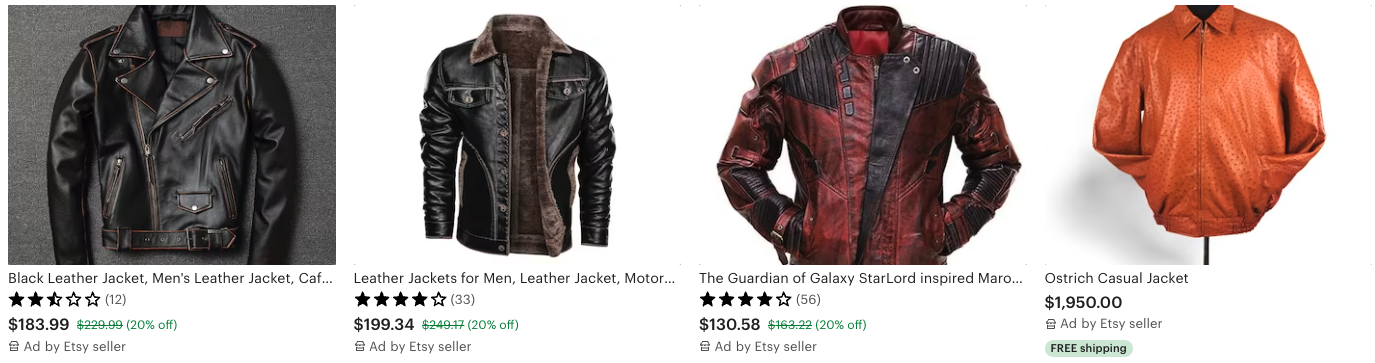} }}%
    \caption{Ad results from the query \textit{jacket} for a user who recently interacted with mens leather jackets.}%
    \label{fig:ad_row_personalized}%
\end{figure}

\section{Introduction}
\label{introduction}

Online advertising is a multi-billion dollar industry (>$\$300$ billion in the fiscal year 2022) \cite{forbes} and sponsored search is one of its largest sub-fields. In the "pay-per-click" model, advertisers are charged for clicks. Thus, sponsored search systems have an incentive to engage users via accurate real-time predictions of clicks and purchases. Deep learning methods have been employed in recent years to improve the performance of these systems in online advertising \cite{zhang2021deep, zhai2016deepintent, zhang2022dhen, zhou2018deep}. More generally, powerful deep learning architectures have been employed to leverage signals from user behaviors and personalize industrial applications, for better user satisfaction and business outcomes \cite{chen2019behavior, xu2022rethinking, pancha2022pinnerformer, zhou2018deep}.

In this article we introduce a general approach to personalization in industrial large-scale applications via a methodology called the adSformer diversifiable personalization module (ADPM). We illustrate how this methodology is deployed to personalize Etsy Ads ranking and bidding systems in order to demonstrate the method's effectiveness. The ADPM-personalized sponsored search system has been deployed to $100\%$ of user traffic and delivering robust online results since February 2023 in Etsy Ads.

We make the following contributions:

\begin{itemize}
\item We introduce the adSformer diversifiable personalization module (ADPM henceforth), a highly configurable module used to encode sequences of recent user actions and derive a short-term dynamic user representation. The module can handle sequences of variable lengths. ADPM's architecture includes three components which learn diverse signals symbiotically: 1. the adSformer encoder component; 2. the pretrained representations component; 3. the learned "on-the-fly" component. We detail ADPM's architecture and explain its implementation for impact and scalability. 

\item The first component of the ADPM, the adSformer encoder, introduces a novel variation on the classic transformer block, where a global max pooling layer is added at the last stage, borrowing from the computer vision literature. We explain the architecture design choice and learning behavior.

\item We demonstrate the module's effectiveness in practice by describing how the ADPM is employed to personalize sponsored search (ad) ranking. We include offline and online experimental results and deployment considerations. We share reproducibility details including optimal hyperparameters, model design choices, pretrained representations workflows, and training approaches.
\end{itemize}

\subsection{Related Work}

 \subsubsection{Personalization from user sequences} A useful survey of deep learning methods employed in sponsored search is \cite{zhang2021deep}. A few other works include sequences of user actions to more generally improve click-through rate prediction models in \cite{zhou2019deep, pi2020search, pi2019practice, chen2019behavior, xu2022rethinking, zhou2018deep, aslanyan2020personalized, supersonalization, pancha2022pinnerformer, kang2018self, grbovic2018real}, which is an essential task in online advertising. Recent developments in personalization emerged to separate sequences into long-term and recent or short-term. Pinterest describes an end-to-end ranking application in \cite{xu2022rethinking}, with a brief reference to recent user actions encoded via a multi-headed self-attention layer and fed to downstream ranking tasks such as search, recommendations, and ads. In personalized eCommerce search, \cite{aslanyan2020personalized} at Ebay had employed the representations of the most recent five clicked items to personalize search rankers.

 \subsubsection{Transformers in personalization} Transformer encoders have been used in personalized product search, both at retrieval and re-ranking phases. The transformer architecture captures the context of users' intents by encoding the sequential nature of user behavior sequences as seen in previous works from Alibaba, Pinterest, and others \cite{chen2019behavior, kang2018self, sun2019bert4rec, pancha2022pinnerformer}. Particularly relevant to our work is the behavioral sequence transformer in \cite{chen2019behavior} which encodes a sequence of last 20 actions to personalize downstream click-through rate models. Focusing on the retrieval phase, TEM, a transformer based model, can tune the amount of personalization applied based on the query characteristics \cite{bi2020transformer}, illustrating the versatility of the transformer architecture. In addition to employing attention mechanisms and transformer architecture, authors in \cite{zhang2020personalized} employ graph neural networks to build a session-aware recommender, where a session represents a short-term user sequence. In a similar vein, sequence and graph modelling came together as a heterogeneous user graph transformer in \cite{shui2022sequence} to improve personalized recommendations.

The rest of the article is organized as follows. We introduce the three-component ADPM in Section \ref{sec:adpm}. Component two of the ADPM employs pretrained representations therefore we detail Etsy Ads' representation learning workflows in Section \ref{pretrained}. To demonstrate ADPM's effectiveness and generality, we describe how this module can be configured and deployed to personalize sponsored search ranking models in Section \ref{sec:ranking}. We present offline ablation studies and online experiment results in Section \ref{sec:experiments}, after which we conclude.

\section{ADPM Methodology}
\label{sec:adpm}

\begin{figure}[tbp]
  \centering
  \includegraphics[scale=0.3]{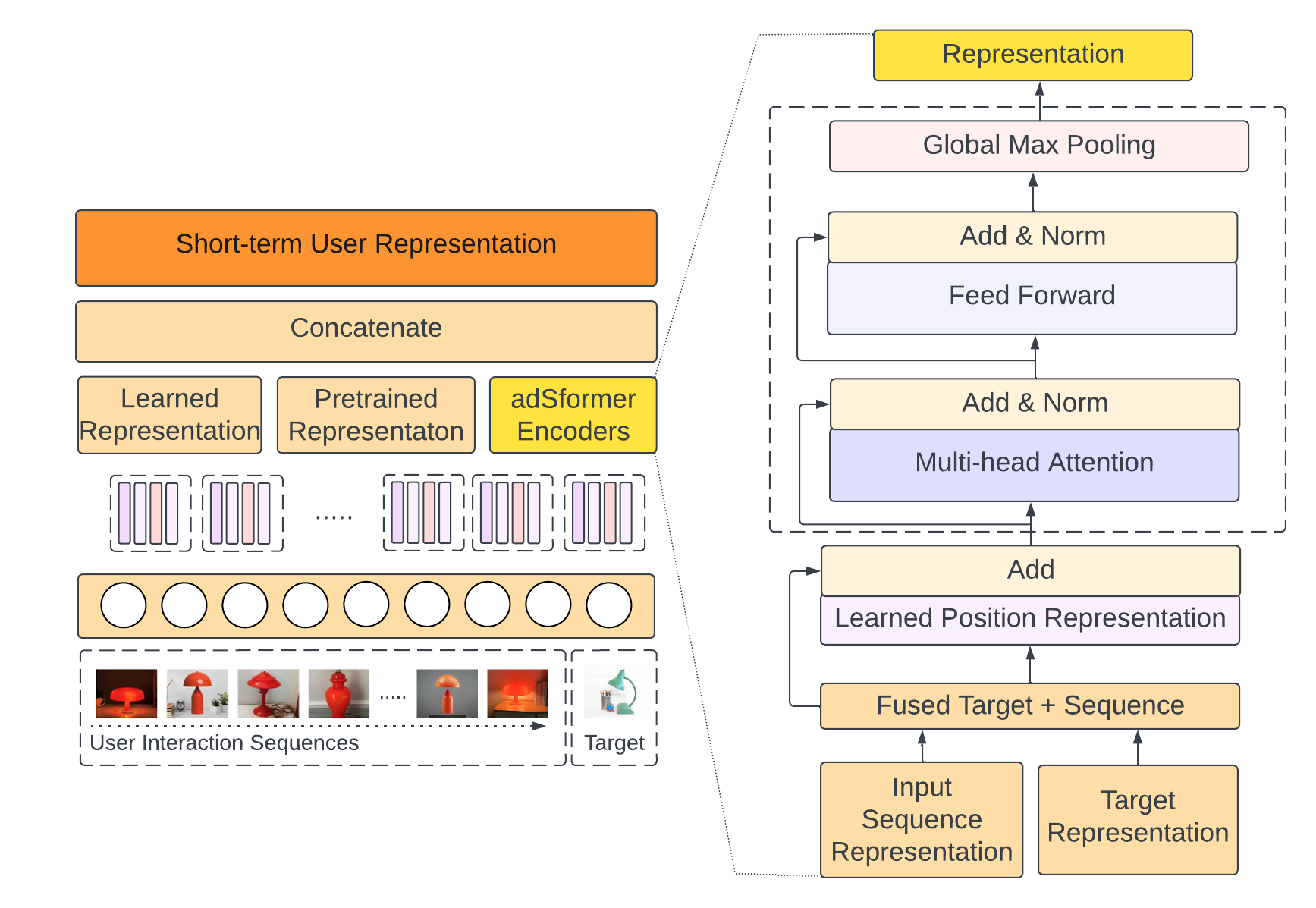}
  \caption{The adSformer diversifiable personalization module (ADPM) derives a dynamic user representation from variable-length sequences of user actions. The adSformer Encoder component includes the adSformer block displayed on the right.}%
  \label{fig:adSFormer_module_and_block}
\end{figure}

\subsection{Variable Length Sequences}

The ADPM, as seen in left-hand side of Figure \ref{fig:adSFormer_module_and_block}, encodes sequences of recent user actions $S_i = (a_i, e_i)$ of variable lengths ${y_i}$, variable entity types ${e_i}$ and variable action types $a_i$. Examples of $(e, a)$ pairs are $e=listingID$ and $a=viewed$. The sequence length $y$ is a random variable taking values in $\{0, ..., M\}$, where $M$ is the maximum sequence length dictated by the engineering system. Realized values of $y$ are dictated by the time delta threshold, which is set to one hour in most of our experiments. All ADPM-encoded user actions would have happened within this time delta, and actions outside the decided upon time delta are masked. The ADPM outputs a short-term dynamic user representation which captures complex relationships within and between different source sequences due to its three components. The ADPM can be used within the context of any model that can benefit from real time personalization in industrial large-scale applications. 

\subsection{Diversity of Signals from ADPM's Components}

The three-component ADPM design is motivated by its intended use in personalization. Diversity of information helps to capture the nuanced patterns of individual users. The three ADPM components learn different signals from the sequences of user actions. This diversity of signals goal is typical when designing neural network architectures and the ADPM architecture achieves this goal for user sequence modelling. The signals learned by each component complement each other which fosters symbiotic learning. In deployed applications, this symbiotic diversity of signals improves robustness to input distribution shifts. The diversity of learned signals also leads to more relevant personalized results through a better derived user representation. We evaluate the quality of this representation in experiments discussed in Section \ref{sec:effectiveness}. Lastly, the diverse signals learned by the three components help the ADPM better adapt to diverse downstream tasks such as ranking, retrieval, and recommendations.

\subsection{Component One: adSformer Encoders}

The adSformer encoder component uses one or more custom adSformer blocks as seen in Figure \ref{fig:adSFormer_module_and_block}, right panel. This component learns a deep, expressive representation of the input sequence. The adSformer block modifies the standard transformer block in \cite{vaswani2017attention} by adding a final global max pooling layer. This layer downsamples the block's outputs by extracting the most salient signals from the sequence representation. 

The adSformer block starts with an embedding layer which encodes listing IDs into dense vectors of size $d_1=32$. We pad the variable length sequence to a common length of $M$ and mask the padding token. We concatenate the target listing representation at position zero as in \cite{chen2019behavior}. We then add a fully learnable position embedding of same dimension $d_1$ to learn the sequence order for the concatenated $ (target, sequence) $ as in \cite{devlin2018bert}. A multi-head self attention layer with scaled dot-product attention \cite{vaswani2017attention} follows. First we get

\begin{equation}
\text{A}(\bQ, \bK, \bV) = \text{softmax}\big(\frac{\bQ\bK^T}{\sqrt{d}}\big)\bV,
\end{equation}

\noindent where $\bQ$, $\bK$ and $\bV$ are query, keys and values respectively, and $A$ is the attention. Then the multi-head self attention is
\begin{equation}
\text{MHSA}(EP) = \text{Concat}(head_1, head_2,\cdots,head_h)\bW^H,
\end{equation}

\noindent with $head_i = \text{A}(\bE\bW^Q, \bE\bW^K, \bE\bW^V)$. The projection matrices are $\bW^Q, \bW^K, \bW^V \in \mathbb{R}^{d \times d}$, and $EP$ represents listing embedding added to position embedding $EP=E+P$. Then the block adds point-wise feed-forward networks $FFN$, $LeakyRelu$ non-linearity as in \cite{chen2019behavior}, and residual connections, dropout and layer normalization in the usual sequence. To simplify notation, $x$ is the output of the previous layer in the block and $g(x)$ represents the next layer:

\begin{align}
g(x) &= x + Dropout(MHSA(x))\\
x &= LayerNorm(x + g(x)) \\
g(x) &= Dropout(FFN(LeakyRelu(x)) \\
x &= LayerNorm(x + g(x)). \\
\end{align}

We add the global max pooling layer at the last stage

\begin{equation}
o_1 = GlobalMaxPooling(x).
\end{equation}

The global pooling layer downsamples the output of the transformer block to a representation vector of size $d_1$ instead of outputting the concatenated transformer block features for the entire sequence. We thus retain the most salient signal, in a parameter efficient manner. We experiment with replacing the global max pooling layer with global average pooling layer and present offline ablation results in Section \ref{sec:experiments}.

\subsection{Component Two: Pretrained Representations}

Component two encodes sequences using listing ID pretrained representations together with average pooling. Depending on downstream performance and availability, we choose from multimodal (AIR) representations detailed in Section \ref{sec:air}, visual representations detailed in Section \ref{sec:visual}, or interaction-based representations detailed in \ref{sec:skip-gram}. Thus component two encodes rich image, text, and multimodal signals from all the listings in the variable length sequence. Formally, for a given sequence of listing IDs' embedding vectors $e_i \in \mathbb{R}^{d_2}$ with $d_2=256$, we compute 

\begin{equation}
o_2 = GlobalAveragePooling([e_1, e_2, ..., e_{y}])
\end{equation}

where $y$ is the sequence length. Then the output of ADPM's component two $o_2$ is a sequence representation vector of size $d_2$. The individual listing ID's pretrained representations are kept frozen in downstream tasks, not requiring an expensive gradient calculation, while adding multimodal signal.

\subsection{Component Three: Representations Learned "On the Fly"} 

The third component of the ADPM introduces representations learned for each sequence from scratch in its own vector space as part of the downstream models. This component learns light weight representations for many different $(e, a)$ sequences for which we do not have pretrained representations. For example we encode sequences of entities $e$ of type \{listingID, taxonomyID, shopID\} for user actions $a$ of type \{cart add, favorited, purchased\}. Formally, for $S_i = (a_i, e_i)$ of variable lengths ${y_i}$, variable entity types ${e_i}$ and variable action types $a_i$, we embed each entity $e_i$ in the action space $a_i$ to get a vector representation $E_{aei} \in \mathbf{R}^{d_{ae}}$ of dimensions $d_{ae}$,

\begin{align}
E_{S_i} &= GlobalAveragePooling([e_1, e_2, ..., e_{y_i}]) \\
o_3 &= Concat([E_{S_1}, E_{S_2}, ..., E_{S_z}]) 
\end{align}

\noindent with $i \in \{1, ..., z\}$ and $z$ is the number of sequences encoded by component three.

\subsection{Dynamic User Representation}

Given an input set of $p$ user action sequences $S_i = (a_i, e_i)$, $i \in \{1, ..., p\}$, of variable lengths $y_i$, the ADPM layer concatenates the three components' outputs to derive a short term dynamic user representation of size $d_1+d_2+\sum d_{ae}$.

\begin{equation}
u = ADPM(S_1, S_2, ..., S_p) = Concat([o_1, o_2, o_3]).
\end{equation}

We concatenate this user representation $u$ to the input layer in downstream personalization tasks.

\begin{figure*}[!tbp]
    \centering
    \includegraphics[scale=0.45]{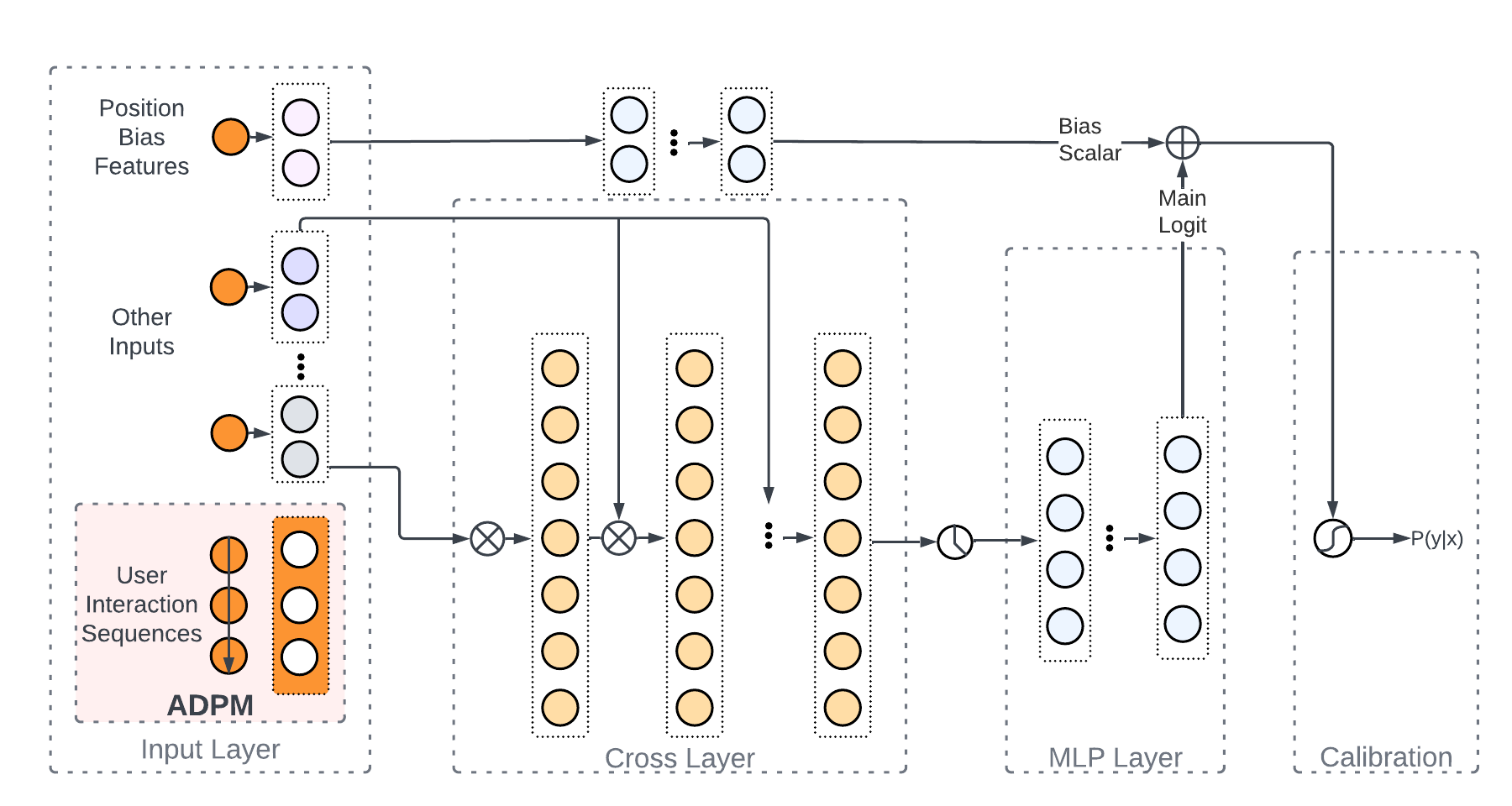}
    \caption{Etsy Ads personalizes sponsored search's CTR and PCCVR models by concatenating the short-term dynamic user representation from ADPM to the input representation layer. Both models share a similar architecture with slight differences in numbers of layers and hidden units.}%
    \label{fig:model_archs}%
\end{figure*}

\subsection{ADPM Implementation}
\label{sec:ADPM}
We designed the ADPM as a general plug-and-play module and implement it as a TensorFlow \cite{abadi2016tensorflow} Keras layer which can be reused across personalization use-cases with a simple import statement. The ADPM layer is configurable, allowing arguments for customizing the ADPM's three components. Since sequences of streamed user actions are made available by the Feature System team to all downstream Etsy applications, ADPM-based personalization can be easily scaled to multiple downstream personalization tasks while at the same time introducing a per-model customization to avoid sacrificing performance.

\subsubsection{Vocabularies} ADPM's sequence representations are implemented as lookup tables. However, when vocabularies are large, the corresponding weight matrices can take up too much memory and disk space, a significant challenge in deployments. To overcome this challenge and reduce memory consumption, we limit the vocabulary size for each sequence to the top $K$ most frequent entities, tuning $K$ as a hyperparameter in downstream CTR and PCCVR models. For example, the CTR's vocabulary corresponding to sequences of listing IDs was capped to top $K=750,000$ listings, down from 100 million in total. The top $K$ listings carry the largest impact, after which increasing vocabulary size led to only marginal improvements in model performance. Hence vocabulary size impact on model performance follows a power distribution \cite{roberts2017expectation}. We built the vocabularies dynamically, from the corpus of variable-length sequences present in the downstream models' training datasets, further maximizing impact. The time delta is a hyperparameter that we can easily tune through ADPM's flexible implementation.

\subsubsection{Software implementation of the pretrained representations component} Pretrained representations are useful for feature encoding, scalability and modularity and we describe learning pretrained representations for every active listing in Section \ref{pretrained}. Running this process offline allows for deeper architectures without latency concerns. The pretrained representation vectors are read by ADPM's component two through the use of lookup tables and a mapping between vocabulary and index positions. To make pretrained representations easily accessible to downstream models, we wrap the lookup table in a TensorFlow SavedModel which includes average pooling layers and handling of default padding values. By saving the lookup table with the downstream model, we assure that the representations used at training time are the same ones used at serving time.

\section{Pretrained Representation Learning}
\label{pretrained}

\begin{figure}[h]
    \begin{minipage}[b]{0.3\linewidth}
        \subfloat[\centering Target listing]{{
            \includegraphics[width=20mm]{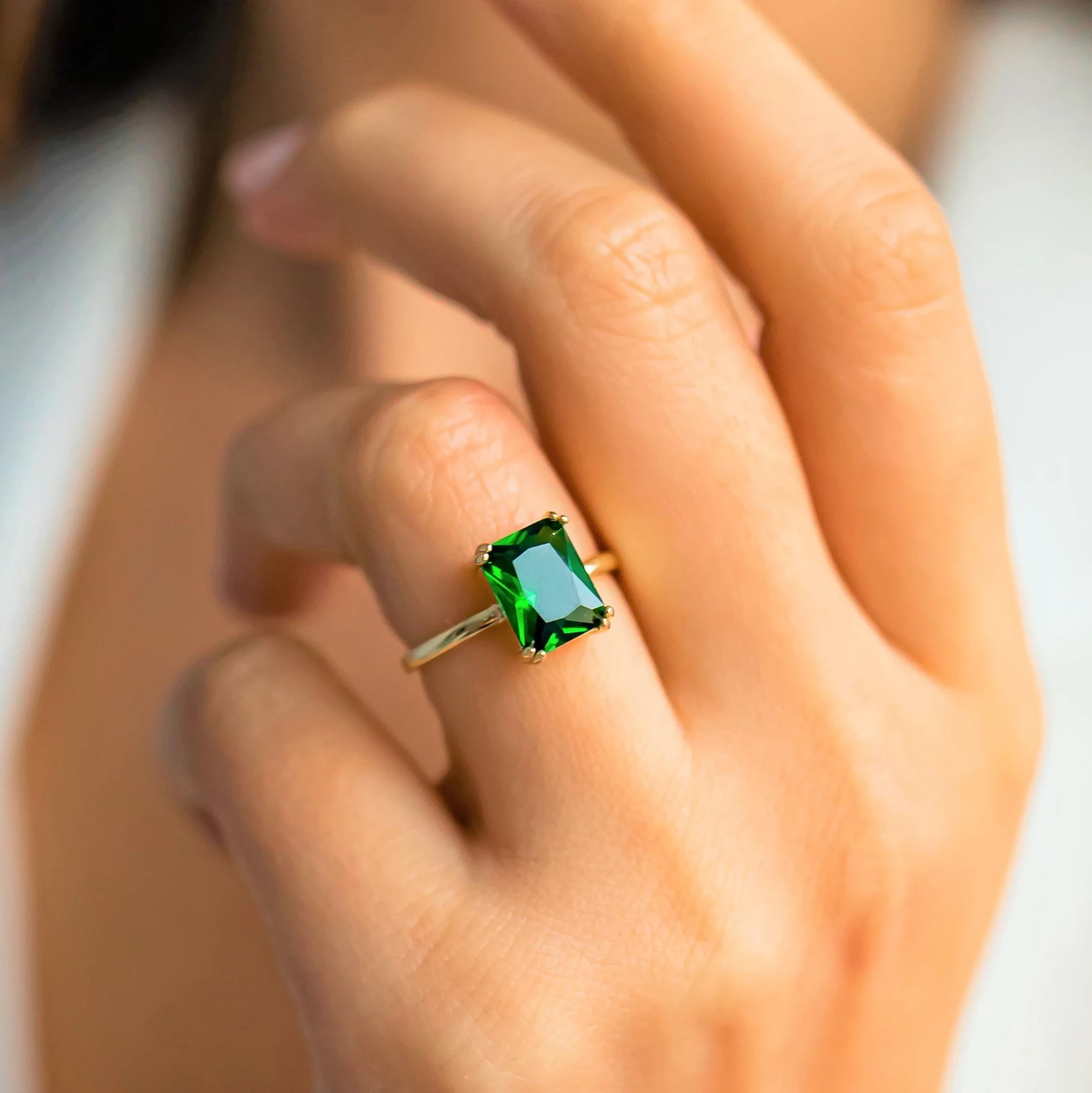}
            \label{fig:emb_target}}}
    \end{minipage}%
    \begin{minipage}[b]{0.7\linewidth}
        \subfloat[\centering AIR representation]{{
            \includegraphics[width=0.85\linewidth]{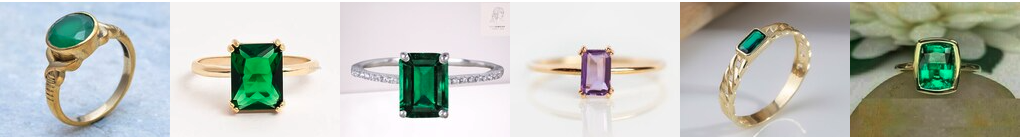}
            \label{fig:emb_air_knn}}}
    
        \subfloat[\centering Visual representation]{{
            \includegraphics[width=0.85\linewidth]{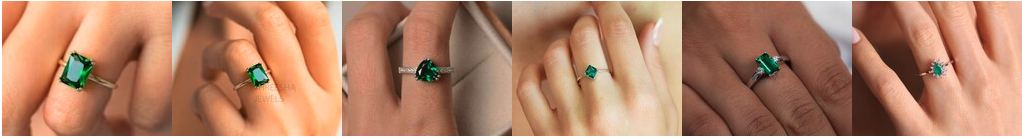}
            \label{fig:emb_visual}}}
    \end{minipage}
    
    \caption{Given a target listing of an emerald ring shown in (a), we visualize the top $6$ nearest neighbors based on the listing's pretrained AIR representation (b) and listing's visual representation (c).}
    \label{fig:emb_knn_comparison}
\end{figure}

 Component two of the ADPM employs pretrained representations to encode variable-length sequences of user actions. For this reason we give an overview of Etsy Ads' pretrained representation learning workflows. Several flavors of representations are trained and used to encode listing IDs: visual, multimodal (AIR), and skip-gram. ADPM's component two's optimum pretrained representation depends on the specifics of the downstream tasks, as illustrated in ablation studies from Section \ref{sec:experiments}. Figure \ref{fig:emb_knn_comparison} shows that the top k nearest neighbors for a listing are different when using visual versus AIR representations. AIR representations better encode price, category, and text while the visual representations better encode color, shape, and background image.

\subsection{Visual Representations} 
\label{sec:visual}
Pinterest authors lament the scarcity of visual representations across large-scale industrial applications \cite{beal2022billion}. In Etsy Ads, we employ image signals across a variety of tasks, such as visually similar candidate generation, search by image, learning other pretrained representations, and the ADPM's second component.

\subsubsection{Model design} We train image representations using a multitask classification architecture \cite{pinterest19unified}, an improvement over the "classic" classification as a proxy to metric learning \cite{pinterest18classification}. By using multiple classification heads, such as taxonomy, color, and material, our representations are able to capture more diverse information about the image. An EfficientNetB0 architecture \cite{efficientnet} with weights pretrained on ImageNet \cite{deng2009imagenet} served as the backbone, and the final layer was replaced with a 256-dimensional convolutional block, the desired output representation size. Image random rotation, translation, zoom, and a color contrast transformation to augment the dataset were applied during training.

\subsubsection{Multitask learning of visual representations} 
\begin{figure}[h]
\includegraphics[width=3.25in]{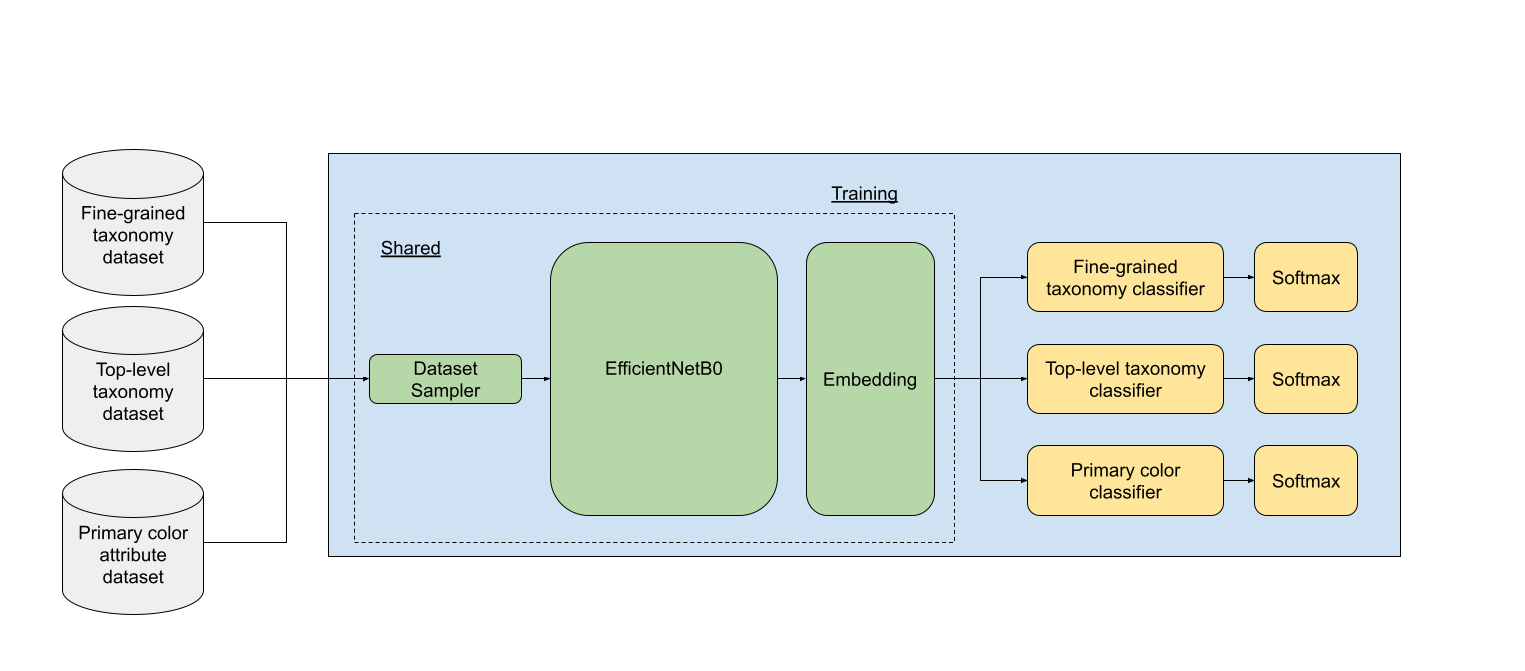}
\centering
\caption{Multitask and multi-dataset visual representations training architecture. The data sampler combines examples from an arbitrary number of datasets corresponding to respective classification heads.}
\label{fig:viz-multitask-arch-3ds}
\end{figure}

Rather than using a single dataset of listing images with multiple label columns, heterogeneous dataset sources containing product images with different selected attributes as the labels were included. Listing attributes such as color and material are optionally input by sellers, so they can be sparse. To mitigate this concern we implemented a dataset sampler, pictured in Figure \ref{fig:viz-multitask-arch-3ds}, which takes in the separate datasets, each with its own unique label type. The sampler evenly distributes examples from each dataset to construct balanced training batches and ascertain that only loss coming from the respective task’s classification head is considered during backpropagation. One additional benefit of this approach is that we can mix in datasets of images from different visual domains to support more use cases, such as user-uploaded review photos. The multitask visual representations are integrated into the visually similar ad recommendations as well as the \textit{Search by Image} application built by Etsy Ads, with more details included in \cite{dolev2023efficient} and Etsy's blog post \cite{EtsyCodeAsCraftCsb1}.

\subsection{Multimodal Representations: Learning AIR}
\label{sec:air}

\subsubsection{Model design} 
\begin{figure}[h]
\includegraphics[width=3.25in]{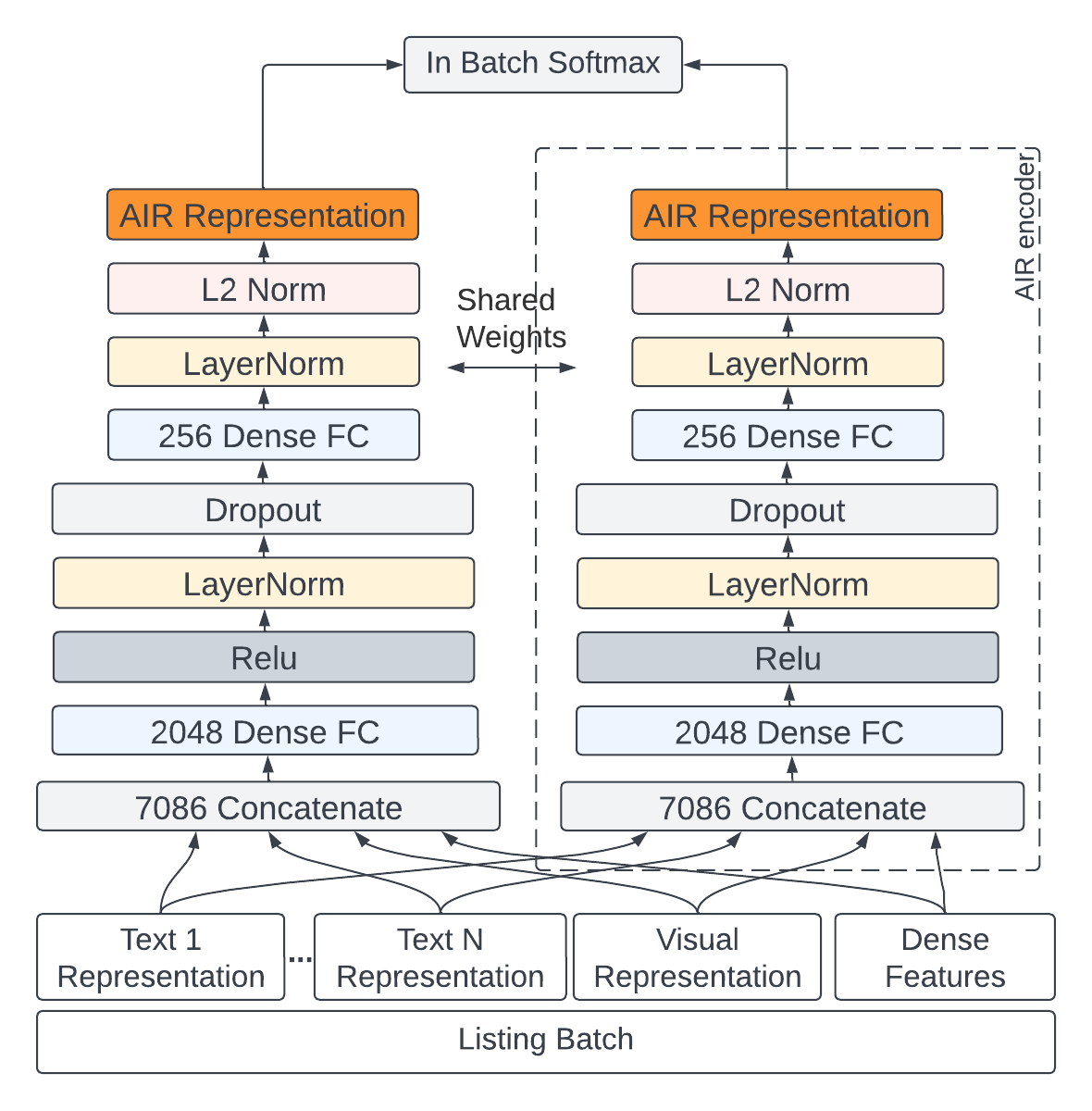}
\centering
\caption{Learning AIR: pseudo-two-tower architecture with shared weights.}
\label{fig:air-encoder}
\end{figure}

We design Ads Information Retrieval (AIR) pretrained listing representations with the explicit goal of driving ad clicks. The AIR model is a neural network with a \textit{pseudo-two-tower} architecture pictured in Figure \ref{fig:air-encoder}. The source and candidate towers share all trainable weights, for which reason we add the prefix "pseudo" to the canonical two-tower architecture \cite{yi2019sampling, yang2020mixed}. 

\subsubsection{Learning} We compile a dataset of listing pairs that have been clicked together on various pages and we consider one listing as the "source" and the other listing as the "candidate" within the same example. The training dataset includes source-candidate pairs, where for each source we consider its corresponding candidate pair as the positive label, while a sample of the other examples in the same batch are the negative labels. For each of the source and candidate listings, we preprocess and concatenate into an input layer a set of multimodal features. For example, we concatenate the multitask visual representations described in Section \ref{sec:visual}, text representations of listing’s title, tags and taxonomy path average pooled from lightweight fastText pretrained representations \cite{bojanowski2017enriching}, and a series of other normalized features. We train the AIR model with a softmax loss \cite{pinterest19unified} and a classification objective. During training, we inference source and candidate 256-dimensional representations for each pair in the batch. We then compute a matrix of cosine similarity scores between each example’s source and candidate representations \cite{nigam2019semantic}. Finally, we compute the classification loss using these scores. 

\subsection{Skip-Gram Listing Representations}
 \label{sec:skip-gram}

Visual and AIR pretrained listing representations do not capture signals from the sequential browsing behavior of users within a web session. Therefore we learn a listing representation from sequences of listings in a browsing session by employing the skip-gram \cite{skipgram} model, similarly to \cite{grbovic2018real, etsy2018iteminteraction}. We learn a vector representation of dimension $d=64$ for each unique listing in the training set. We use a hierarchical softmax loss function (\cite{morin2005hierarchical} and \cite{mnih2008scalable}), which showed much better results than classic negative sampling. We found that fastText library \footnote{https://fasttext.cc/} worked well for training, as long as we disabled functionality to consider subwords. 

\section{ADPM - Personalized Sponsored Search Ranking}
\label{sec:ranking}

\subsection{Background}

\begin{figure}[bp]
  \centering
  \includegraphics[scale=0.4]{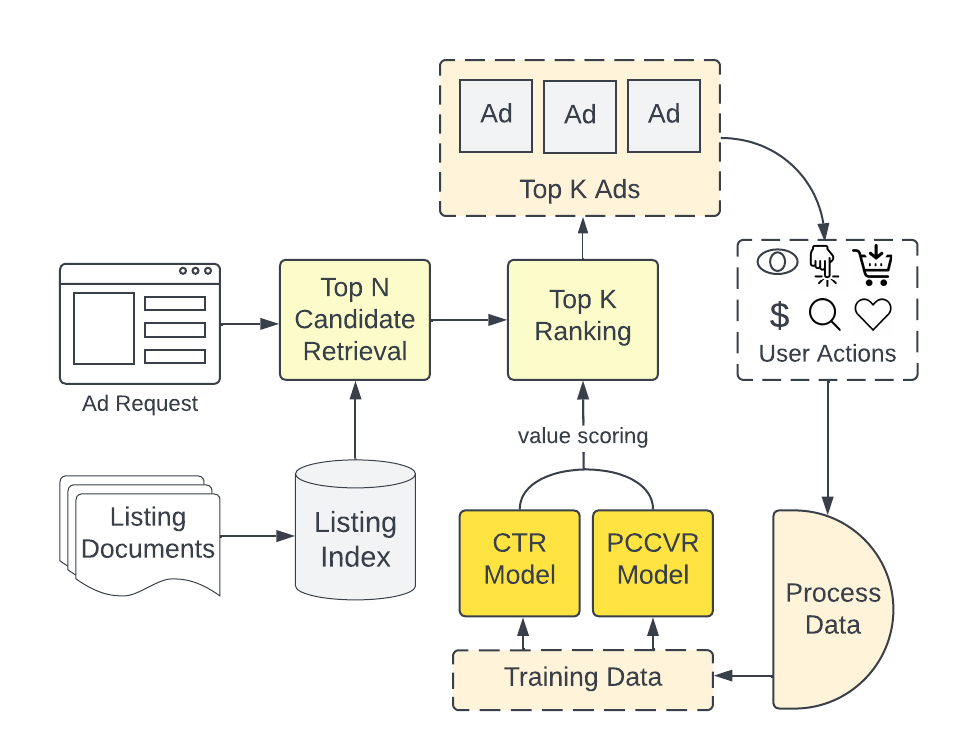}
  \caption{In our Learning to Rank system, ads are ranked by a value score.}
  \label{fig:ltr}
\end{figure}

 To demonstrate ADPM's effectiveness and generality, we describe how Etsy Ads employes it to personalize ranking and bidding models in sponsored search. First, we'll give an overview of the sponsored search system. As many online advertising services, sellers sponsor their listings (ads) through a second-price cost-per-click auction campaign. In order to decide which ads to show to a user, Etsy Ads employs Learning to Rank (LTR) \cite{liu2011learning, cao2007learning} framework in Figure \ref{fig:ltr}. The system is divided according to the classic two-stage dichotomy: \textit{ads candidate retrieval} and then \textit{ads ranking}. First, the candidate retrieval step winnows the full inventory from over 100 million listings (items) to under 1000 candidates, after which results are reranked based on a combined value score obtained from the ranking and bidding models predictions, the CTR and PCCVR models respectively.

In the CTR case, $p(\mathbf{x})$ denotes the predicted probability $p(y_{CTR}=1)$ that the candidate listing will be clicked. For PCCVR $p(\mathbf{x})$ denotes the predicted conditional probability $p(y_{PCCVR}=1|y_{CTR}=1)$ that the listing will be purchased, having been clicked \cite{EtsyCodeAsCraftCsb}.  

 Before including the ADPM, baseline production models' architectures in Etsy Ads were standard stacked deep and cross neural networks (DCN) \cite{wang2021dcn}. We used these as baselines in all our offline and online comparisons. The CTR's DCN has four cross and four deep layers with sizes ${5000, 2500, 250, 500}$ and PCCVR's DCN employs two cross and two deep layers of sizes 240, and 120 respectively.

 \subsection{ADPM-Personalized CTR and PCCVR}

The goal is to personalize CTR and PCCVR using the ADPM for improved user outcomes. We employ the ADPM to encode sequences of recent user actions anywhere on Etsy for both logged-in and logged-out users. Instead of considering the last M user actions as typically seen in \cite{pancha2022pinnerformer, chen2019behavior, xu2022rethinking}, we consider a sliding one hour window of user actions, in reverse chronological order of timestamps, to encode only recent behavior. More formally, let $S = \left( a, e, t \right)$, be a one-hour sequence of user actions, where $a$ is the action type one of \{\text{view}, \text{favorite}, \text{cart add}, \text{purchase}, \text{search}\}, and $e$ represents one of the entities \{listingID, shopID, categoryID, text query\} associated with action $a$ performed at timestamp $t$. Due to semantics and infrastructure constraints we further cap the maximum sequence length at $M=50$ actions. Each sequence $I$ is truncated to within one hour of the most recent action, so $t_0-t_{last} \leq 1 hr$. The resulting sequences have variable length which the ADPM can handle through padding and masking.

 We concatenate ADPM's output, the dynamic user representation, to the rest of the wide input feature vector \cite{cheng2016wide} similarly to \cite{chen2019behavior}. The optimum ADPM configuration for the three components differs between the CTR and the PCCVR models as Table \ref{tab:configs} details, and this flexibility is one of ADPM's strengths. Each configuration choice is similar to a hyperparameter that we tune.

\begin{table}
  \caption{The different CTR and PCCVR ADPM configurations}
  \label{tab:configs}
  \footnotesize
  \begin{tabular}{lrrrr}
    \toprule
    Config & CTR & PCCVR\\
    \midrule
    Pretrained Representation Type & AIR & skip\&visual \\
    Num adSformer Blocks & 1 & 1 \\
    adSformer Block Dropout & 0 & 0 \\
    adSformer Num Heads & 3 & 2 \\
    Learned Listing Representation Dimension & 32 & 32\\
    Learned Shop Representation Dimension & 16 & 16  \\
    Learned Taxonomy Representation Dimension & 8 & 8 \\
    (e, a) (Learned Component) & (all e, all a) & (all e, all a) \\
    (e, a)((Pretrained Component) & (listing, all a) & (listing, all a) \\
    (e, a) (adSformer Encoder) & (listing, viewed) & (listing, viewed) \\
    (listing, viewed) Vocabulary Size & 750K & 650K\\
     Num OOV Tokens & 1 & 1 \\
  \bottomrule
\end{tabular}
\end{table}

After employing ADPM personalization we call the new models the adSformer CTR and PCCVR. Figure \ref{fig:model_archs} depicts the model architecture corresponding to the adSformer CTR and PCCVR after ADPM inclusion.

\subsubsection{The adSformer CTR}

 For the  adSformer CTR model, the adSformer encoder component includes one adSformer block with three attention heads. The adSformer encoder encodes user and browser sequences of recently viewed listings since these $(e, a)$ have the highest session frequency. Within the pretrained representations component, the multimodal pretrained representation (AIR) described in Section \ref{sec:air} works best for the CTR task to encode all sequences of listing IDs. ADPM's third, "on the fly", component encodes all one-hour sequences of recent user actions. 

After concatenating the ADPM's output to the input representation layer, it is critical to have a cross network (DCN) in the personalized adSformer CTR architecture \cite{wang2017deep}. As recognized in recent work \cite{zhang2022dhen, lang2021architecture, song2019autoint, li2020interpretable, wang2021dcn, roberts2020qr, zhang2022dhen}, learning higher order feature interactions effectively from the input layer helps performance in large scale CTR prediction. For the adSformer CTR, the interaction module (DCN) is necessary to fully leverage the wide input representation which includes the ADPM. The cross layer exclusion leads to a 1.17\% drop in ROC AUC. The large capacity of the adSformer CTR model aids learning and generalization by providing a smoother loss landscape and more stable and faster learning dynamics. The adSformer CTR only needs one epoch of training, for a total of 11 hours (one A100 GPU) using Adam \cite{kingma2014adam} optimizer. Throughout model training we decay the learning rate using cosine annealing \cite{loshchilov2016sgdr}. We select the largest batch size that can fit in memory (8192) and tune the learning rate to an optimum $lr=0.002$.

\subsubsection{The adSformer PCCVR}

To obtain the ADPM-personalized adSformer PCCVR model, we configure the ADPM and concatenate its output user representation to the baseline PCCVR model input layer, similarly to the CTR. Table \ref{tab:configs} gives the ADPM configuration for the adSformer PCCVR model. The major differences to the CTR's case is a smaller optimum adSformer encoder component with only two heads. The PCCVR trains on smaller datasets. We parameterize the ADPM's pretrained representations component with pretrained skip-gram and visual listing representations. While we trained the PCCVR baseline production model on only two weeks of historical click logs (135 million examples), the adSformer PCCVR has more parameters after the ADPM addition and thrives with three weeks of data (200mln examples).

\subsubsection{Learning}
Both the CTR and PCCVR before and after ADPM-based personalization are formulated as binary classification problems, and we use a binary cross entropy loss function 
\begin{equation}
L = -\frac{1}{N} \sum_{\left( \mathbf{x}, y \right ) \in D} \left( ylogp \left( \mathbf{x} \right) + \left( 1-y \right) log \left( 1-p \left(\mathbf{x}\right)\right)\right),
\end{equation}
where $D$ represent all samples in the training dataset, $y \in \{ 0, 1 \}$ is the label, $p(x)$ is the predicted probability, and $x$ is the input vector containing (listing, query, and context) attributes as well as the ADPM's user representation output.

\section{Experimental Results}
\label{sec:experiments}

We evaluate ADPM's effectiveness through offline and online experiments by comparing the adSformer CTR and adSformer PCCVR models to their non-personalized baselines described in Section \ref{sec:ranking}. We also perform ablation studies where we compare the ADPM with other user sequence modeling approaches. Finally, we compare the effectiveness of permutations of ADPM configurations in offline experiments.

\subsection{Offline Experiments}

We evaluate offline performance using PR-AUC and ROC-AUC metrics and present lifts as compared to baselines in all the ablation studies.

\subsubsection{Evaluating The Effectiveness of ADPM's Components and Their Symbiotic Learning Behavior}
\label{sec:effectiveness}

We aim to better understand the symbiotic learning behavior from the combination of the three components of the ADPM. Our hypothesis is that each component learns different signals and together through this diversity lead to better outcomes in downstream tasks. For a comparison, we permute the configuration of the ADPM holding constant all other modelling choices and provide lifts in the CTR and PCCVR AUC metrics. 

Furthermore we compare the three-component ADPM to a baseline encoder of user sequences, such as the BST employed by \cite{chen2019behavior}, which uses an eight head transformer encoder with one block to encode a sequence of 20 user actions and their timestamp. Other alternatives exist, such as the attention mechanisms in \cite{zhou2018deep} or \cite{xu2022rethinking} but the BST model in \cite{chen2019behavior} outperforms most of these approaches according to \cite{chen2019behavior}. We also compare against a simple average embedding aggregation of the last five actions, similarly to Ebay's \cite{aslanyan2020personalized}. The goal is to understand if the ADPM is more effective through its design choices then these baselines when used to personalize a downstream task, such as the CTR model. 

We strive to make comparisons as relevant as possible by using the same underlying sequence lengths and datasets, training budgets, as well as hyperparameters from the original papers, however some differences remain dictated by the implementation pipelines. For example, we do learn the position embedding for the BST similarly to the ADPM instead of using timestamp deltas but do not foresee major differences as noted in \cite{vaswani2017attention}. When employed to personalize the CTR model in ADPM's place, the BST runs out of memory for a one hour sequence and eight heads (as in the BST article) at an embedding size of $32$ and for the same batch size employed in all experiments, so we downsize to five and three heads. 

One of ADPM's strength is the ability to derive maximum signal with constraints on training and serving resources. We also proxy Ebay's \cite{aslanyan2020personalized} encoding of last $k=5$ user actions as an alternative baseline for ADPM. In the original article BST is run on a fixed sequence length of 20, so we apply both the BST and the ADPM with three components on a sequence of most recent 20 actions as well. To summarize the results in Tables \ref{tab:adpm_permute_ctr} and \ref{tab:adpm_permute_PCCVR}, the three-component ADPM outperforms all the other component combinations as well as the BST and other baselines, although the BST is a strong alternative.

\begin{table}[h]
    \centering
    \footnotesize
     \caption{Ablation studies showing offline lift in metrics within the CTR downstream task. Each variant is compared against the baseline, non-personalized, CTR model. Each variant represents either a configuration of the three component ADPM or another personalization method we compare the ADPM against. We vary the ADPM's three components: Comp1=adSformerEncoder; Comp2=Pretrained; Comp3=On-the-fly. Personalization baselines are: BST=Alibaba's Behavioral Sequence Transformer; Ebay's last k=5.}
    \begin{tabular}{lrrrr}
        \hline
        & \multicolumn{2}{c}{CTR} \\ 
        Configuration & 
              ROC-AUC & 
              PR-AUC \\
        \midrule
        $ADPM-3[MaxPool]$ &
              \textbf{+2.71}\% & 
              \textbf{+8.88}\% \\
         $ADPM-3[AvgPool]$ &
              +2.67\% & 
              +8.36\% \\
        $[Comp1, Comp2]$ & 
              +2.42\% & 
              +7.19\% \\
        $[Comp1, Comp3]$ &
              +2.04\% & 
              +6.36\% \\
        $[Comp2, Comp3]$ & 
              +2.43\% & 
              +7.65\% \\
        $[Comp1]$ & 
              +1.71\% & 
              +5.46\% \\
        $[Comp2]$ & 
              +2.09\% & 
              +6.31\% \\
        $[Comp3]$ & 
              +1.75\% & 
              +5.74\% \\
        $[ADPM-k=20]$ & 
              +2.66\% & 
              +8.5\% \\
        $[BST-k=20-8heads]$ & 
              +1.77\% & 
              +5.55\% \\
        $[BST-1-hr-5heads]$ & 
              +1.82\% & 
              +3.99\%  \\
        $[BST-1-hr-3heads]$ & 
              +1.78\% & 
              +4.31\% \\
        $[Ebay-k=5]$ &
            +1.65\% &
            +4.58\% \\
         $[ADPM-k=5]$ &
            +2.43\% &
            +7.4\% \\
         $[BST-k=5]$ &
            +1.45\% &
            +4.77\% \\
        \bottomrule
    \end{tabular}
    \label{tab:adpm_permute_ctr}
\end{table}

\begin{table}[h]
    \centering
    \footnotesize
     \caption{Ablation studies comparing downstream performance for permutations of ADPM's components within the PCCVR downstream task.}
    \begin{tabular}{lrrrr}
        \hline
        & \multicolumn{2}{c}{PCCVR} \\ 
        Configuration & 
              ROC-AUC & 
              PR-AUC \\
        \midrule
        $ADPM-3[MaxPool]$ &
              \textbf{+2.42}\% & 
              \textbf{+28.47}\% \\
         $ADPM-3[AvgPool]$ &
              +2.37\% &
              +26.76\% \\
        $[Comp1, Comp2]$ & 
              +1.09\% &
              +13.15\% \\
        $[Comp1, Comp3]$ &
              +1.87\% &
              +18.14\% \\
        $[Comp2, Comp3]$ & 
              +2.12\% &
              +24.38\% \\
        $[Comp1]$ & 
              +0.36\% &
              +3.47\% \\
        $[Comp2]$ & 
              +0.84\% &
              +10.76\% \\
        $[Comp3]$ & 
              +1.55\% &
              +15.48\% \\
        \bottomrule
    \end{tabular}
    \label{tab:adpm_permute_PCCVR}
\end{table}

We have additionally experimented with a an average pooling layer replacing the final global max pooling layer in the adSformer block within ADPM's component one. Results are similar for the two, which makes intuitive sense since the average is influenced by extreme values in a sample such as the max. The global max pooling is still more effective in deriving a different signal from the sequence which complements the signals derived by the other two components. Interestingly, ADPM also outperformes when encoding the last five or the last 20 instead of a one-hour, variable-length, sequence.

\subsubsection{Evaluating the Impact of Different Choices of Pretrained Representations in ADPM's Component Two}

 In Table \ref{tab:emb_eval_comparison} we can see how different configurations of the pretrained representation component two of ADPM lead to different test ROC AUC and PR AUC. We notice that combining all three flavors of pretrained representations in CTR's ADPM configuration leads to the highest test ROC AUC. However only AIR was included in the deployed adSformer CTR, since both ROC AUC and PR AUC are top ranking and memory requirements are lower. In PCCVR's ADPM configuration skip-gram and visual representations performed best together in terms of PR AUC, however the ADPM configuration with only the adSformer encoder component and the learned component performed well. One hypothesis is that the adSformer encoder learns a signal similar to the signal from the pretrained skip-gram since they both encode the sequential nature of sequences of listing IDs.
 
\begin{table}[h]
    \centering
    \footnotesize
     \caption{Ablation studies showing offline lift in metrics against the baseline, non-personalized, models for various configurations of the ADPM's pretrained representation component.}
    \begin{tabular}{lrrrr}
        \hline
        & \multicolumn{2}{c}{CTR} & \multicolumn{2}{c}{PCCVR} \\ 
        Pretrained Selection & 
              ROC-AUC & 
              PR-AUC & 
              ROC-AUC & 
              PR-AUC \\ 
        \midrule
        Skipgram &
              +1.99\% & 
              +5.88\% & 
              +1.35\% &
              +9.93\% \\
        Visual & 
              +1.90\% & 
              +5.37\% & 
              +1.33\% &
              +10.21\% \\
        AIR &
              +2.14\% & 
              \textbf{+6.44\%} & 
              +1.23\% &
              +8.19\% \\
        Skipgram/Visual & 
              +2.10\% & 
              +5.71\% & 
              +1.32\% &
              \textbf{+10.38\%}\\
        AIR/Skipgram & 
              +2.20\% & 
              +6.44\% & 
              +1.25\% &
              +7.51\% \\
        AIR/Visual & 
              +2.16\% & 
              +6.22\% & 
              +1.17\% &
              +8.54\% \\
        AIR/Skipgram/Visual & 
              \textbf{+2.26\%} & 
              +6.17\% & 
              +1.14\% &
              +7.56\% \\
        No Pretrained Representations & 
              +1.60\% & 
              +4.89\% & 
              \textbf{+1.37\%} &
              +10.24\% \\
        \bottomrule
    \end{tabular}
    \label{tab:emb_eval_comparison}
\end{table}

 The adSformer CTR and PCCVR differ in ADPM's pretrained component optimal configuration. We are not surprised, given that a user may have different intents during their browsing journey. For example, a buyer shown an ad impression is likely to be earlier in the purchase funnel and the click intent (CTR prediction) may be around shopping for inspiration or price comparisons, so the AIR embedding works better. Post-click the purchase intent (PCCVR prediction) may be more narrowly focused on stylistic variations and shipping profiles of a candidate listing, so the visual signal from the image embedding is more important. 

 For reproducibility, Table \ref{tab:configs1} gives hyperparameters and other machine learning choices for the adSformer CTR as well as for its baseline, the non-personalized standard CTR prediction model. The adSformer PCCVR has similar settings.

\begin{table}
  \caption{Hyperparameters for training the adSformer CTR and its baseline, the non-personalized CTR prediction model which is a neural network with four fully connected layers.}
  \label{tab:configs1}
  \tiny
  \begin{tabular}{lrrrr}
    \toprule
    Config & adSformer CTR & baseline NN CTR\\
    \midrule  
    Num Epochs & 1  & 3 \\
    Learning Rate & 0.002  & 0.00061\\
    Learning Rate Scheduler & cosine & cosine\\
    Optimizer & Adam & LazyAdam\\
    Adam $\beta_1$ & 0.9 & 0.9 \\
    Adam $\beta_2$ & 0.999 & 0.999 \\
    Adam $\epsilon$ & 1e-08 & 1e-08\\
    Dropout Hidden Layers & 0 & 0\\
    Batch Norm Hidden Layer& yes & yes \\
    Cross Module (DCN) & yes & no \\
    Batch Size (train) & 8192 & 8192\\
    Batch Size (validation) & 500 & 500\\
    Machine Type (training) & A100 & P100\\
    Machine Type (evaluation) & A100 & T4\\
    Loss & BinaryCrossEntropy  & BinaryCrossEntropy\\
    Num Parameters (trainable) & 708,504,031 & 233,709,623\\
    Num Parameters (total) & 897,499,741 & 243,726,397 \\
  \bottomrule
    \end{tabular}
\end{table}

Table \ref{tab:configs_data} gives the dataset choices for the winning adSformer CTR and PCCVR. For the CTR we sample 50 percent of non-clicked impressions in the training dataset but perform a random sampling for the validation set, so that validation and test sets are representative of the production data seen in the wild. The PCCVR is trained on clicked impressions but evaluated on a representative test sample, similarly to the adSformer CTR.

\begin{table}
  \caption{Train and validation datasets characteristics for adSformer CTR and PCCVR}
  \label{tab:configs_data}
  \footnotesize
  \begin{tabular}{lrrrr}
    \toprule
    Model & Config & Training & Validation \\
    \midrule  
    CTR & Num Examples & ~300mln & ~8.5mln \\
    CTR & Num Consecutive Days & 30 & 1 \\
    CTR & Sampling Type & 50-50 negative sampling & random sampling \\
    PCCVR & Num Examples & ~200mln & ~8.5mln \\
    PCCVR & Num Consecutive Days & 21 & 1\\
    PCCVR & Sampling Type & click filtered & random sampling \\
  \bottomrule
    \end{tabular}
\end{table}

\begin{figure}[h]
\includegraphics[width=3.25in]{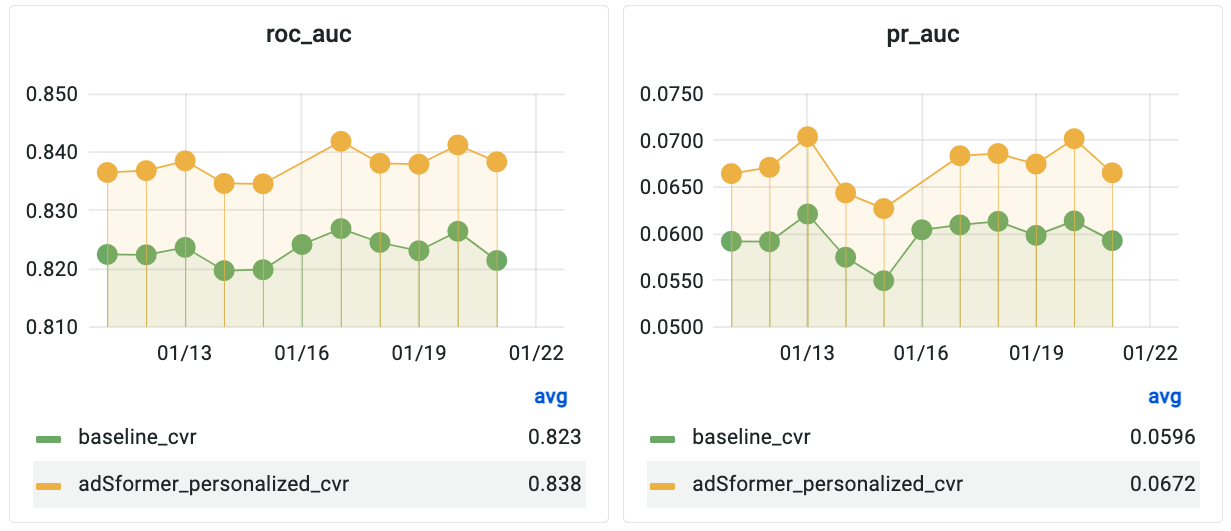}
\centering
\caption{Daily offline AUC evaluation of adSformer personalized PCCVR model against the baseline PCCVR model from 2023-01-11 to 2023-01-21.}
\label{fig:cvr-offline-eval}
\end{figure}

In Etsy Ads production, models are trained daily and evaluated on next-day test datasets. In Figure \ref{fig:cvr-offline-eval} we can see the PR AUC and ROC AUC lift for the ADPM-personalized adSformer PCCVR as compared to the non-personalized PCCVR production baseline across a period of ten days.

\subsection{Online A/B Experiments}

Before deploying to $100\%$ traffic we evaluated the adSformer CTR and adSformer PCCVR in online A/B tests with results in Table \ref{tab:ads-exp}. Compared to the previous production model, the adSformer PCCVR demonstrated an increase in Return on Ad Spend (ROAS), PCCVR, and Gross Merchandised Sale (GMS). For Etsy Ads sellers this translates to an increase in orders with larger sale values at a lower cost. The CTR model shows improved engagement metrics and the adSformer CTR and PCCVR models deployed in tandem for A/B test demonstrated even larger increases in ROAS and GMS when comparing to individual model contributions.

\begin{table}
    \centering
    \caption{Statistically significant online A/B experiment results (p-value < 0.001).}
    \label{tab:ads-exp}
    \begin{tabular}{lrrrr}
        \toprule
        Metric & ADPM-CTR & ADPM-PCCVR  & Tandem \\
        \cmidrule(r){1-1}\cmidrule(l){2-4}
        Ads CTR & +2.55\% & -0.02\% & +2.42\% \\
        Ads PCCVR & +0.29\% & +1.26\%  & +0.90\%\\
        Ads ROAS & +1.02\% & +3.51\% & +5.34\%\\
        Ads GMS & +3.02\% & +2.24\% & +4.17\% \\
        \bottomrule
    \end{tabular}
\end{table}

\subsection{Additional Considerations in the Deployed ADPM-Personalized Ads System}

Sampling biases, such as position bias \cite{huang2021deep}, can reduce ad ranking improvements from ADPM. A correlation can be observed between the positional rank of a shown ad and the probability of being clicked, which can lead to a feedback loop where the model inadvertently prioritizes certain candidates due to their previous high ranking and not due to model improvement. To address this challenge, an auxiliary debiasing model \cite{zhao2019recommending} was included in the adSformer CTR. 

Predicted CTR and PCCVR scores must reflect true probabilities in order to assign monetary value and forecast budget depletion, therefore a calibration layer is also introduced for each of the CTR and PCCVR models. Calibration is achieved through the parametric approach of Platt Scaling \cite{platt1999probabilistic, guo2017calibration}.

Numerous real world machine learning applications across natural language processing \cite{radford2019language}, computer vision \cite{beal2022billion}, trajectory prediction \cite{roberts2019neural}, personalized ads, search and recommendations \cite{zhao2019recommending, baltescu2022itemsage}, and computational advertising \cite{ma2018entire, ma2018modeling} employ multitask learning \cite{caruana1997multitask}. A technique many decades old, multitask learning commonly improves accuracy of tasks as well as efficiency of machine learning pipelines. However, the case for multitask learning in mature machine learning systems that have been deployed as separate models for many years, is complex in practice. For example, models such as the adSformer CTR and the adSformer PCCVR have been tuned separately with different choices in terms of training datasets, neural network capacities, and optimum ADPM configurations. Etsy Ads is in the process of experimenting with the multitask learning paradigm in personalized sponsored search.

\subsection{Infrastructure Considerations in Deployed ADPM-Personalized Ads System}

Introducing most recent / real time sequences into a production training and serving path includes infrastructure challenges, and a heavy investment in streaming architecture. Thankfully, much of this infrastructure has already been standardized across Etsy. Etsy's Feature System team uses a low latency streaming system powered by Kafka PubSub for writing interaction event logs to a low latency feature store. The logged-in and logged-out user features are made available through a scalable micro service. The fetched in-session features are captured and preserved in training logs as close as possible to model serving, which minimizes skew between training and serving feature set. Our model serving system uses memcache to cache model scores thereby reducing latency and volume to downstream feature fetching and model inference services. The key used to write and read requests must be updated to include user and browser ID (for logged-out users). The real-time nature of the user action sequences demands that either the cache is removed or that the expiration TTL is lowered dramatically. At scale, in our production systems, we pay close attention to latency and cloud cost surges due to increased volume to downstream services.

\section{Conclusion}

 In this article we introduced a scalable general approach to ads personalization from variable-lengthsequences of recent user actions, the adSformer diversifiable personalization module (ADPM). Following the success of deploying the ADPM-personalized sponsored search, Etsy Ads is currently experimenting with ADPM-personalized recommendation ads. Future work includes further experimenting with the ADPM's three components by creating other adSformer encoder architectures, adding pretrained graph representations to component two, and updating image, text, and multimodal representations. To this end, Etsy Ads continues improving their pretrained representations through strengthening the ads computer vision capabilities. Finally, Etsy Ads plans to extend artificial intelligence driven personalization efforts to the seller toolbox via generative image, multimodal, and video technologies.

\section{Acknowledgments}
We thank engineers in Etsy Ads for their work leading to developing and deploying the adSformers. In alphabetical order: Julie Chien, Eric Herrmann, Julia Hoffman, Rezwana Karim, Mason Kirchner, Melissa Louie, Alec Malstrom, Seoyoon Park, Maharshi Thakker, Justin Tse, Gerardo Veltri. We also thank the Rivulet streaming features team, Information Retrieval Platform, and the ML Platform team for their support.

\bibliographystyle{ACM-Reference-Format}
\bibliography{bibliography}


\begin{thebibliography}{63}


\ifx \showCODEN    \undefined \def \showCODEN     #1{\unskip}     \fi
\ifx \showDOI      \undefined \def \showDOI       #1{#1}\fi
\ifx \showISBNx    \undefined \def \showISBNx     #1{\unskip}     \fi
\ifx \showISBNxiii \undefined \def \showISBNxiii  #1{\unskip}     \fi
\ifx \showISSN     \undefined \def \showISSN      #1{\unskip}     \fi
\ifx \showLCCN     \undefined \def \showLCCN      #1{\unskip}     \fi
\ifx \shownote     \undefined \def \shownote      #1{#1}          \fi
\ifx \showarticletitle \undefined \def \showarticletitle #1{#1}   \fi
\ifx \showURL      \undefined \def \showURL       {\relax}        \fi
\providecommand\bibfield[2]{#2}
\providecommand\bibinfo[2]{#2}
\providecommand\natexlab[1]{#1}
\providecommand\showeprint[2][]{arXiv:#2}

\bibitem[for(2023)]%
        {forbes}
 \bibinfo{year}{2023}\natexlab{}.
\newblock \bibinfo{booktitle}{\emph{Forecast 2023: Ad Spending Will Slow Down
  Next Year But Will Continue To Grow}}.
\newblock
\urldef\tempurl%
\url{https://www.forbes.com/sites/bradadgate/2022/12/08/forecast-2023-ad-spending-will-slow-down-next-year-but-will-continue-to-grow/}
\showURL{%
\tempurl}


\bibitem[Abadi et~al\mbox{.}(2016)]%
        {abadi2016tensorflow}
\bibfield{author}{\bibinfo{person}{Mart{\'\i}n Abadi}, \bibinfo{person}{Paul
  Barham}, \bibinfo{person}{Jianmin Chen}, \bibinfo{person}{Zhifeng Chen},
  \bibinfo{person}{Andy Davis}, \bibinfo{person}{Jeffrey Dean},
  \bibinfo{person}{Matthieu Devin}, \bibinfo{person}{Sanjay Ghemawat},
  \bibinfo{person}{Geoffrey Irving}, \bibinfo{person}{Michael Isard},
  {et~al\mbox{.}}} \bibinfo{year}{2016}\natexlab{}.
\newblock \showarticletitle{TensorFlow: A system for large-scale machine
  learning}. In \bibinfo{booktitle}{\emph{12th USENIX symposium on operating
  systems design and implementation (OSDI 16)}}. \bibinfo{pages}{265--283}.
\newblock


\bibitem[Aslanyan et~al\mbox{.}(2020)]%
        {aslanyan2020personalized}
\bibfield{author}{\bibinfo{person}{Grigor Aslanyan}, \bibinfo{person}{Aritra
  Mandal}, \bibinfo{person}{Prathyusha Senthil~Kumar}, \bibinfo{person}{Amit
  Jaiswal}, {and} \bibinfo{person}{Manojkumar Rangasamy~Kannadasan}.}
  \bibinfo{year}{2020}\natexlab{}.
\newblock \showarticletitle{Personalized ranking in eCommerce search}. In
  \bibinfo{booktitle}{\emph{Companion Proceedings of the Web Conference 2020}}.
  \bibinfo{pages}{96--97}.
\newblock


\bibitem[Awad et~al\mbox{.}(2022)]%
        {EtsyCodeAsCraftCsb}
\bibfield{author}{\bibinfo{person}{Alaa Awad}, \bibinfo{person}{Congzhe Su},
  {and} \bibinfo{person}{Erica Greene}.} \bibinfo{year}{2022}\natexlab{}.
\newblock \bibinfo{booktitle}{\emph{How We Built A Context-Specific Bidding
  System for Etsy Ads}}.
\newblock
\urldef\tempurl%
\url{https://www.etsy.com/codeascraft/how-we-built-a-context-specific-bidding-system-for-etsy-ads}
\showURL{%
\tempurl}


\bibitem[Baltescu et~al\mbox{.}(2022)]%
        {baltescu2022itemsage}
\bibfield{author}{\bibinfo{person}{Paul Baltescu}, \bibinfo{person}{Haoyu
  Chen}, \bibinfo{person}{Nikil Pancha}, \bibinfo{person}{Andrew Zhai},
  \bibinfo{person}{Jure Leskovec}, {and} \bibinfo{person}{Charles Rosenberg}.}
  \bibinfo{year}{2022}\natexlab{}.
\newblock \showarticletitle{{ItemSage}: Learning product embeddings for
  shopping recommendations at {Pinterest}}.
\newblock \bibinfo{journal}{\emph{arXiv preprint arXiv:2205.11728}}
  (\bibinfo{year}{2022}).
\newblock


\bibitem[Beal et~al\mbox{.}(2022)]%
        {beal2022billion}
\bibfield{author}{\bibinfo{person}{Josh Beal}, \bibinfo{person}{Hao-Yu Wu},
  \bibinfo{person}{Dong~Huk Park}, \bibinfo{person}{Andrew Zhai}, {and}
  \bibinfo{person}{Dmitry Kislyuk}.} \bibinfo{year}{2022}\natexlab{}.
\newblock \showarticletitle{Billion-scale pretraining with vision transformers
  for multi-task visual representations}. In
  \bibinfo{booktitle}{\emph{Proceedings of the IEEE/CVF Winter Conference on
  Applications of Computer Vision}}. \bibinfo{pages}{564--573}.
\newblock


\bibitem[Bi et~al\mbox{.}(2020)]%
        {bi2020transformer}
\bibfield{author}{\bibinfo{person}{Keping Bi}, \bibinfo{person}{Qingyao Ai},
  {and} \bibinfo{person}{W~Bruce Croft}.} \bibinfo{year}{2020}\natexlab{}.
\newblock \showarticletitle{A transformer-based embedding model for
  personalized product search}. In \bibinfo{booktitle}{\emph{Proceedings of the
  43rd International ACM SIGIR Conference on Research and Development in
  Information Retrieval}}. \bibinfo{pages}{1521--1524}.
\newblock


\bibitem[Bojanowski et~al\mbox{.}(2017)]%
        {bojanowski2017enriching}
\bibfield{author}{\bibinfo{person}{Piotr Bojanowski}, \bibinfo{person}{Edouard
  Grave}, \bibinfo{person}{Armand Joulin}, {and} \bibinfo{person}{Tomas
  Mikolov}.} \bibinfo{year}{2017}\natexlab{}.
\newblock \showarticletitle{Enriching word vectors with subword information}.
\newblock \bibinfo{journal}{\emph{Transactions of the association for
  computational linguistics}}  \bibinfo{volume}{5} (\bibinfo{year}{2017}),
  \bibinfo{pages}{135--146}.
\newblock


\bibitem[Cao et~al\mbox{.}(2007)]%
        {cao2007learning}
\bibfield{author}{\bibinfo{person}{Zhe Cao}, \bibinfo{person}{Tao Qin},
  \bibinfo{person}{Tie-Yan Liu}, \bibinfo{person}{Ming-Feng Tsai}, {and}
  \bibinfo{person}{Hang Li}.} \bibinfo{year}{2007}\natexlab{}.
\newblock \showarticletitle{Learning to rank: from pairwise approach to
  listwise approach}. In \bibinfo{booktitle}{\emph{Proceedings of the 24th
  international conference on Machine learning}}. \bibinfo{pages}{129--136}.
\newblock


\bibitem[Caruana(1997)]%
        {caruana1997multitask}
\bibfield{author}{\bibinfo{person}{Rich Caruana}.}
  \bibinfo{year}{1997}\natexlab{}.
\newblock \showarticletitle{Multitask learning}.
\newblock \bibinfo{journal}{\emph{Machine learning}} \bibinfo{volume}{28},
  \bibinfo{number}{1} (\bibinfo{year}{1997}), \bibinfo{pages}{41--75}.
\newblock


\bibitem[Chen et~al\mbox{.}(2019)]%
        {chen2019behavior}
\bibfield{author}{\bibinfo{person}{Qiwei Chen}, \bibinfo{person}{Huan Zhao},
  \bibinfo{person}{Wei Li}, \bibinfo{person}{Pipei Huang}, {and}
  \bibinfo{person}{Wenwu Ou}.} \bibinfo{year}{2019}\natexlab{}.
\newblock \showarticletitle{Behavior sequence transformer for e-commerce
  recommendation in {A}libaba}. In \bibinfo{booktitle}{\emph{Proceedings of the
  1st International Workshop on Deep Learning Practice for High-Dimensional
  Sparse Data}}. \bibinfo{pages}{1--4}.
\newblock


\bibitem[Cheng et~al\mbox{.}(2016)]%
        {cheng2016wide}
\bibfield{author}{\bibinfo{person}{Heng-Tze Cheng}, \bibinfo{person}{Levent
  Koc}, \bibinfo{person}{Jeremiah Harmsen}, \bibinfo{person}{Tal Shaked},
  \bibinfo{person}{Tushar Chandra}, \bibinfo{person}{Hrishi Aradhye},
  \bibinfo{person}{Glen Anderson}, \bibinfo{person}{Greg Corrado},
  \bibinfo{person}{Wei Chai}, \bibinfo{person}{Mustafa Ispir}, {et~al\mbox{.}}}
  \bibinfo{year}{2016}\natexlab{}.
\newblock \showarticletitle{Wide \& deep learning for recommender systems}. In
  \bibinfo{booktitle}{\emph{Proceedings of the 1st workshop on deep learning
  for recommender systems}}. \bibinfo{pages}{7--10}.
\newblock


\bibitem[Comet.com(2021)]%
        {CometML}
\bibfield{author}{\bibinfo{person}{Comet.com}.}
  \bibinfo{year}{2021}\natexlab{}.
\newblock \bibinfo{booktitle}{\emph{{Comet.com} home page}}.
\newblock
\urldef\tempurl%
\url{https://www.comet.com/}
\showURL{%
\tempurl}


\bibitem[Deng et~al\mbox{.}(2009)]%
        {deng2009imagenet}
\bibfield{author}{\bibinfo{person}{Jia Deng}, \bibinfo{person}{Wei Dong},
  \bibinfo{person}{Richard Socher}, \bibinfo{person}{Li-Jia Li},
  \bibinfo{person}{Kai Li}, {and} \bibinfo{person}{Li Fei-Fei}.}
  \bibinfo{year}{2009}\natexlab{}.
\newblock \showarticletitle{Imagenet: A large-scale hierarchical image
  database}. In \bibinfo{booktitle}{\emph{2009 IEEE conference on computer
  vision and pattern recognition}}. Ieee, \bibinfo{pages}{248--255}.
\newblock


\bibitem[Devlin et~al\mbox{.}(2018)]%
        {devlin2018bert}
\bibfield{author}{\bibinfo{person}{Jacob Devlin}, \bibinfo{person}{Ming-Wei
  Chang}, \bibinfo{person}{Kenton Lee}, {and} \bibinfo{person}{Kristina
  Toutanova}.} \bibinfo{year}{2018}\natexlab{}.
\newblock \showarticletitle{Bert: Pre-training of deep bidirectional
  transformers for language understanding}.
\newblock \bibinfo{journal}{\emph{arXiv preprint arXiv:1810.04805}}
  (\bibinfo{year}{2018}).
\newblock


\bibitem[Dolev and Awad(2023)]%
        {EtsyCodeAsCraftCsb1}
\bibfield{author}{\bibinfo{person}{Eden Dolev} {and} \bibinfo{person}{Alaa
  Awad}.} \bibinfo{year}{2023}\natexlab{}.
\newblock \bibinfo{booktitle}{\emph{From Image Classification to Multitask
  Modeling: Building Etsy’s Search by Image Feature}}.
\newblock
\urldef\tempurl%
\url{https://www.etsy.com/codeascraft/from-image-classification-to-multitask-modeling-building-etsys-search-by-image-feature}
\showURL{%
\tempurl}


\bibitem[Dolev et~al\mbox{.}(2023)]%
        {dolev2023efficient}
\bibfield{author}{\bibinfo{person}{Eden Dolev}, \bibinfo{person}{Alaa Awad},
  \bibinfo{person}{Denisa Roberts}, \bibinfo{person}{Zahra Ebrahimzadeh},
  \bibinfo{person}{Marcin Mejran}, \bibinfo{person}{Vaibhav Malpani}, {and}
  \bibinfo{person}{Mahir Yavuz}.} \bibinfo{year}{2023}\natexlab{}.
\newblock \showarticletitle{Efficient Large-Scale Vision Representation
  Learning}.
\newblock \bibinfo{journal}{\emph{arXiv preprint arXiv:2305.13399}}
  (\bibinfo{year}{2023}).
\newblock


\bibitem[Grbovic and Cheng(2018)]%
        {grbovic2018real}
\bibfield{author}{\bibinfo{person}{Mihajlo Grbovic} {and}
  \bibinfo{person}{Haibin Cheng}.} \bibinfo{year}{2018}\natexlab{}.
\newblock \showarticletitle{Real-time personalization using embeddings for
  search ranking at {Airbnb}}. In \bibinfo{booktitle}{\emph{Proceedings of the
  24th ACM SIGKDD International Conference on Knowledge Discovery \& Data
  Mining}}. \bibinfo{pages}{311--320}.
\newblock


\bibitem[Guo et~al\mbox{.}(2017)]%
        {guo2017calibration}
\bibfield{author}{\bibinfo{person}{Chuan Guo}, \bibinfo{person}{Geoff Pleiss},
  \bibinfo{person}{Yu Sun}, {and} \bibinfo{person}{Kilian~Q Weinberger}.}
  \bibinfo{year}{2017}\natexlab{}.
\newblock \showarticletitle{On calibration of modern neural networks}. In
  \bibinfo{booktitle}{\emph{International conference on machine learning}}.
  PMLR, \bibinfo{pages}{1321--1330}.
\newblock


\bibitem[Huang et~al\mbox{.}(2021)]%
        {huang2021deep}
\bibfield{author}{\bibinfo{person}{Jianqiang Huang}, \bibinfo{person}{Ke Hu},
  \bibinfo{person}{Qingtao Tang}, \bibinfo{person}{Mingjian Chen},
  \bibinfo{person}{Yi Qi}, \bibinfo{person}{Jia Cheng}, {and}
  \bibinfo{person}{Jun Lei}.} \bibinfo{year}{2021}\natexlab{}.
\newblock \showarticletitle{Deep position-wise interaction network for CTR
  Prediction}. In \bibinfo{booktitle}{\emph{Proceedings of the 44th
  International ACM SIGIR Conference on Research and Development in Information
  Retrieval}}. \bibinfo{pages}{1885--1889}.
\newblock


\bibitem[Kang and McAuley(2018)]%
        {kang2018self}
\bibfield{author}{\bibinfo{person}{Wang-Cheng Kang} {and}
  \bibinfo{person}{Julian McAuley}.} \bibinfo{year}{2018}\natexlab{}.
\newblock \showarticletitle{Self-attentive sequential recommendation}. In
  \bibinfo{booktitle}{\emph{2018 IEEE international conference on data mining
  (ICDM)}}. IEEE, \bibinfo{pages}{197--206}.
\newblock


\bibitem[Kingma and Ba(2014)]%
        {kingma2014adam}
\bibfield{author}{\bibinfo{person}{Diederik~P Kingma} {and}
  \bibinfo{person}{Jimmy Ba}.} \bibinfo{year}{2014}\natexlab{}.
\newblock \showarticletitle{Adam: A method for stochastic optimization}.
\newblock \bibinfo{journal}{\emph{arXiv preprint arXiv:1412.6980}}
  (\bibinfo{year}{2014}).
\newblock


\bibitem[Lang et~al\mbox{.}(2021)]%
        {lang2021architecture}
\bibfield{author}{\bibinfo{person}{Lang Lang}, \bibinfo{person}{Zhenlong Zhu},
  \bibinfo{person}{Xuanye Liu}, \bibinfo{person}{Jianxin Zhao},
  \bibinfo{person}{Jixing Xu}, {and} \bibinfo{person}{Minghui Shan}.}
  \bibinfo{year}{2021}\natexlab{}.
\newblock \showarticletitle{Architecture and operation adaptive network for
  online recommendations}. In \bibinfo{booktitle}{\emph{Proceedings of the 27th
  ACM SIGKDD Conference on Knowledge Discovery \& Data Mining}}.
  \bibinfo{pages}{3139--3149}.
\newblock


\bibitem[Li et~al\mbox{.}(2020)]%
        {li2020interpretable}
\bibfield{author}{\bibinfo{person}{Zeyu Li}, \bibinfo{person}{Wei Cheng},
  \bibinfo{person}{Yang Chen}, \bibinfo{person}{Haifeng Chen}, {and}
  \bibinfo{person}{Wei Wang}.} \bibinfo{year}{2020}\natexlab{}.
\newblock \showarticletitle{Interpretable click-through rate prediction through
  hierarchical attention}. In \bibinfo{booktitle}{\emph{Proceedings of the 13th
  International Conference on Web Search and Data Mining}}.
  \bibinfo{pages}{313--321}.
\newblock


\bibitem[Liu(2011)]%
        {liu2011learning}
\bibfield{author}{\bibinfo{person}{Tie-Yan Liu}.}
  \bibinfo{year}{2011}\natexlab{}.
\newblock \bibinfo{booktitle}{\emph{Learning to rank for information
  retrieval}}.
\newblock \bibinfo{publisher}{Springer Science \& Business Media}.
\newblock


\bibitem[Loshchilov and Hutter(2016)]%
        {loshchilov2016sgdr}
\bibfield{author}{\bibinfo{person}{Ilya Loshchilov} {and}
  \bibinfo{person}{Frank Hutter}.} \bibinfo{year}{2016}\natexlab{}.
\newblock \showarticletitle{Sgdr: Stochastic gradient descent with warm
  restarts}.
\newblock \bibinfo{journal}{\emph{arXiv preprint arXiv:1608.03983}}
  (\bibinfo{year}{2016}).
\newblock


\bibitem[Ma et~al\mbox{.}(2018b)]%
        {ma2018modeling}
\bibfield{author}{\bibinfo{person}{Jiaqi Ma}, \bibinfo{person}{Zhe Zhao},
  \bibinfo{person}{Xinyang Yi}, \bibinfo{person}{Jilin Chen},
  \bibinfo{person}{Lichan Hong}, {and} \bibinfo{person}{Ed~H Chi}.}
  \bibinfo{year}{2018}\natexlab{b}.
\newblock \showarticletitle{Modeling task relationships in multi-task learning
  with multi-gate mixture-of-experts}. In \bibinfo{booktitle}{\emph{Proceedings
  of the 24th ACM SIGKDD international conference on knowledge discovery \&
  data mining}}. \bibinfo{pages}{1930--1939}.
\newblock


\bibitem[Ma et~al\mbox{.}(2018a)]%
        {ma2018entire}
\bibfield{author}{\bibinfo{person}{Xiao Ma}, \bibinfo{person}{Liqin Zhao},
  \bibinfo{person}{Guan Huang}, \bibinfo{person}{Zhi Wang},
  \bibinfo{person}{Zelin Hu}, \bibinfo{person}{Xiaoqiang Zhu}, {and}
  \bibinfo{person}{Kun Gai}.} \bibinfo{year}{2018}\natexlab{a}.
\newblock \showarticletitle{Entire space multi-task model: An effective
  approach for estimating post-click conversion rate}. In
  \bibinfo{booktitle}{\emph{The 41st International ACM SIGIR Conference on
  Research \& Development in Information Retrieval}}.
  \bibinfo{pages}{1137--1140}.
\newblock


\bibitem[Mikolov et~al\mbox{.}(2013)]%
        {skipgram}
\bibfield{author}{\bibinfo{person}{Tomas Mikolov}, \bibinfo{person}{Kai Chen},
  \bibinfo{person}{Greg Corrado}, {and} \bibinfo{person}{Jeffrey Dean}.}
  \bibinfo{year}{2013}\natexlab{}.
\newblock \showarticletitle{Efficient estimation of word representations in
  vector space}.
\newblock \bibinfo{journal}{\emph{arXiv preprint arXiv:1301.3781}}
  (\bibinfo{year}{2013}).
\newblock


\bibitem[Mnih and Hinton(2008)]%
        {mnih2008scalable}
\bibfield{author}{\bibinfo{person}{Andriy Mnih} {and}
  \bibinfo{person}{Geoffrey~E Hinton}.} \bibinfo{year}{2008}\natexlab{}.
\newblock \showarticletitle{A scalable hierarchical distributed language
  model}.
\newblock \bibinfo{journal}{\emph{Advances in neural information processing
  systems}}  \bibinfo{volume}{21} (\bibinfo{year}{2008}).
\newblock


\bibitem[Morin and Bengio(2005)]%
        {morin2005hierarchical}
\bibfield{author}{\bibinfo{person}{Frederic Morin} {and}
  \bibinfo{person}{Yoshua Bengio}.} \bibinfo{year}{2005}\natexlab{}.
\newblock \showarticletitle{Hierarchical probabilistic neural network language
  model}. In \bibinfo{booktitle}{\emph{International workshop on artificial
  intelligence and statistics}}. PMLR, \bibinfo{pages}{246--252}.
\newblock


\bibitem[Nigam et~al\mbox{.}(2019a)]%
        {nigam2019semantic}
\bibfield{author}{\bibinfo{person}{Priyanka Nigam}, \bibinfo{person}{Yiwei
  Song}, \bibinfo{person}{Vijai Mohan}, \bibinfo{person}{Vihan Lakshman},
  \bibinfo{person}{Weitian Ding}, \bibinfo{person}{Ankit Shingavi},
  \bibinfo{person}{Choon~Hui Teo}, \bibinfo{person}{Hao Gu}, {and}
  \bibinfo{person}{Bing Yin}.} \bibinfo{year}{2019}\natexlab{a}.
\newblock \showarticletitle{Semantic product search}. In
  \bibinfo{booktitle}{\emph{Proceedings of the 25th ACM SIGKDD International
  Conference on Knowledge Discovery \& Data Mining}}.
  \bibinfo{pages}{2876--2885}.
\newblock


\bibitem[Nigam et~al\mbox{.}(2019b)]%
        {amazonNIR}
\bibfield{author}{\bibinfo{person}{Priyanka Nigam}, \bibinfo{person}{Yiwei
  Song}, \bibinfo{person}{Vijai Mohan}, \bibinfo{person}{Vihan Lakshman},
  \bibinfo{person}{Weitian Ding}, \bibinfo{person}{Ankit Shingavi},
  \bibinfo{person}{Choon~Hui Teo}, \bibinfo{person}{Hao Gu}, {and}
  \bibinfo{person}{Bing Yin}.} \bibinfo{year}{2019}\natexlab{b}.
\newblock \showarticletitle{Semantic product search}. In
  \bibinfo{booktitle}{\emph{Proceedings of the 25th ACM SIGKDD International
  Conference on Knowledge Discovery \& Data Mining}}.
  \bibinfo{pages}{2876--2885}.
\newblock


\bibitem[Pancha et~al\mbox{.}(2022)]%
        {pancha2022pinnerformer}
\bibfield{author}{\bibinfo{person}{Nikil Pancha}, \bibinfo{person}{Andrew
  Zhai}, \bibinfo{person}{Jure Leskovec}, {and} \bibinfo{person}{Charles
  Rosenberg}.} \bibinfo{year}{2022}\natexlab{}.
\newblock \showarticletitle{{PinnerFormer}: Sequence modeling for user
  representation at {Pinterest}}.
\newblock \bibinfo{journal}{\emph{arXiv preprint arXiv:2205.04507}}
  (\bibinfo{year}{2022}).
\newblock


\bibitem[Pi et~al\mbox{.}(2019)]%
        {pi2019practice}
\bibfield{author}{\bibinfo{person}{Qi Pi}, \bibinfo{person}{Weijie Bian},
  \bibinfo{person}{Guorui Zhou}, \bibinfo{person}{Xiaoqiang Zhu}, {and}
  \bibinfo{person}{Kun Gai}.} \bibinfo{year}{2019}\natexlab{}.
\newblock \showarticletitle{Practice on long sequential user behavior modeling
  for click-through rate prediction}. In \bibinfo{booktitle}{\emph{Proceedings
  of the 25th ACM SIGKDD International Conference on Knowledge Discovery \&
  Data Mining}}. \bibinfo{pages}{2671--2679}.
\newblock


\bibitem[Pi et~al\mbox{.}(2020)]%
        {pi2020search}
\bibfield{author}{\bibinfo{person}{Qi Pi}, \bibinfo{person}{Guorui Zhou},
  \bibinfo{person}{Yujing Zhang}, \bibinfo{person}{Zhe Wang},
  \bibinfo{person}{Lejian Ren}, \bibinfo{person}{Ying Fan},
  \bibinfo{person}{Xiaoqiang Zhu}, {and} \bibinfo{person}{Kun Gai}.}
  \bibinfo{year}{2020}\natexlab{}.
\newblock \showarticletitle{Search-based user interest modeling with lifelong
  sequential behavior data for click-through rate prediction}. In
  \bibinfo{booktitle}{\emph{Proceedings of the 29th ACM International
  Conference on Information \& Knowledge Management}}.
  \bibinfo{pages}{2685--2692}.
\newblock


\bibitem[Platt et~al\mbox{.}(1999)]%
        {platt1999probabilistic}
\bibfield{author}{\bibinfo{person}{John Platt} {et~al\mbox{.}}}
  \bibinfo{year}{1999}\natexlab{}.
\newblock \showarticletitle{Probabilistic outputs for support vector machines
  and comparisons to regularized likelihood methods}.
\newblock \bibinfo{journal}{\emph{Advances in large margin classifiers}}
  \bibinfo{volume}{10}, \bibinfo{number}{3} (\bibinfo{year}{1999}),
  \bibinfo{pages}{61--74}.
\newblock


\bibitem[Radford et~al\mbox{.}(2019)]%
        {radford2019language}
\bibfield{author}{\bibinfo{person}{Alec Radford}, \bibinfo{person}{Jeffrey Wu},
  \bibinfo{person}{Rewon Child}, \bibinfo{person}{David Luan},
  \bibinfo{person}{Dario Amodei}, \bibinfo{person}{Ilya Sutskever},
  {et~al\mbox{.}}} \bibinfo{year}{2019}\natexlab{}.
\newblock \showarticletitle{Language models are unsupervised multitask
  learners}.
\newblock \bibinfo{journal}{\emph{OpenAI blog}} \bibinfo{volume}{1},
  \bibinfo{number}{8} (\bibinfo{year}{2019}), \bibinfo{pages}{9}.
\newblock


\bibitem[Roberts(2019)]%
        {roberts2019neural}
\bibfield{author}{\bibinfo{person}{Denisa Roberts}.}
  \bibinfo{year}{2019}\natexlab{}.
\newblock \showarticletitle{Neural networks for Lorenz map prediction: A trip
  through time}.
\newblock \bibinfo{journal}{\emph{arXiv preprint arXiv:1903.07768}}
  (\bibinfo{year}{2019}).
\newblock


\bibitem[Roberts and Roberts(2020)]%
        {roberts2020qr}
\bibfield{author}{\bibinfo{person}{Denisa~AO Roberts} {and}
  \bibinfo{person}{Lucas~R Roberts}.} \bibinfo{year}{2020}\natexlab{}.
\newblock \showarticletitle{QR and LQ Decomposition Matrix Backpropagation
  Algorithms for Square, Wide, and Deep--Real or Complex--Matrices and Their
  Software Implementation}.
\newblock \bibinfo{journal}{\emph{arXiv preprint arXiv:2009.10071}}
  (\bibinfo{year}{2020}).
\newblock


\bibitem[Roberts and Roberts(2017)]%
        {roberts2017expectation}
\bibfield{author}{\bibinfo{person}{Lucas Roberts} {and} \bibinfo{person}{Denisa
  Roberts}.} \bibinfo{year}{2017}\natexlab{}.
\newblock \showarticletitle{An expectation maximization framework for
  Yule-Simon preferential attachment models}.
\newblock \bibinfo{journal}{\emph{arXiv preprint arXiv:1710.08511}}
  (\bibinfo{year}{2017}).
\newblock


\bibitem[Shui et~al\mbox{.}(2022)]%
        {shui2022sequence}
\bibfield{author}{\bibinfo{person}{Zeren Shui}, \bibinfo{person}{Ge Liu},
  \bibinfo{person}{Anoop Deoras}, {and} \bibinfo{person}{George Karypis}.}
  \bibinfo{year}{2022}\natexlab{}.
\newblock \showarticletitle{Sequence-graph duality: Unifying user modeling with
  self-attention for sequential recommendation}.
\newblock  (\bibinfo{year}{2022}).
\newblock


\bibitem[Song et~al\mbox{.}(2019)]%
        {song2019autoint}
\bibfield{author}{\bibinfo{person}{Weiping Song}, \bibinfo{person}{Chence Shi},
  \bibinfo{person}{Zhiping Xiao}, \bibinfo{person}{Zhijian Duan},
  \bibinfo{person}{Yewen Xu}, \bibinfo{person}{Ming Zhang}, {and}
  \bibinfo{person}{Jian Tang}.} \bibinfo{year}{2019}\natexlab{}.
\newblock \showarticletitle{Autoint: Automatic feature interaction learning via
  self-attentive neural networks}. In \bibinfo{booktitle}{\emph{Proceedings of
  the 28th ACM International Conference on Information and Knowledge
  Management}}. \bibinfo{pages}{1161--1170}.
\newblock


\bibitem[Sun et~al\mbox{.}(2019)]%
        {sun2019bert4rec}
\bibfield{author}{\bibinfo{person}{Fei Sun}, \bibinfo{person}{Jun Liu},
  \bibinfo{person}{Jian Wu}, \bibinfo{person}{Changhua Pei},
  \bibinfo{person}{Xiao Lin}, \bibinfo{person}{Wenwu Ou}, {and}
  \bibinfo{person}{Peng Jiang}.} \bibinfo{year}{2019}\natexlab{}.
\newblock \showarticletitle{{BERT4Rec}: Sequential recommendation with
  bidirectional encoder representations from transformer}. In
  \bibinfo{booktitle}{\emph{Proceedings of the 28th ACM international
  conference on information and knowledge management}}.
  \bibinfo{pages}{1441--1450}.
\newblock


\bibitem[Tan and Le(2019)]%
        {efficientnet}
\bibfield{author}{\bibinfo{person}{Mingxing Tan} {and} \bibinfo{person}{Quoc~V.
  Le}.} \bibinfo{year}{2019}\natexlab{}.
\newblock \showarticletitle{EfficientNet: rethinking model scaling for
  convolutional neural networks}.
\newblock \bibinfo{journal}{\emph{CoRR}}  \bibinfo{volume}{abs/1905.11946}
  (\bibinfo{year}{2019}).
\newblock
\showeprint[arXiv]{1905.11946}
\urldef\tempurl%
\url{http://arxiv.org/abs/1905.11946}
\showURL{%
\tempurl}


\bibitem[Vaswani et~al\mbox{.}(2017)]%
        {vaswani2017attention}
\bibfield{author}{\bibinfo{person}{Ashish Vaswani}, \bibinfo{person}{Noam
  Shazeer}, \bibinfo{person}{Niki Parmar}, \bibinfo{person}{Jakob Uszkoreit},
  \bibinfo{person}{Llion Jones}, \bibinfo{person}{Aidan~N Gomez},
  \bibinfo{person}{{\L}ukasz Kaiser}, {and} \bibinfo{person}{Illia
  Polosukhin}.} \bibinfo{year}{2017}\natexlab{}.
\newblock \showarticletitle{Attention is all you need}.
\newblock \bibinfo{journal}{\emph{Advances in neural information processing
  systems}}  \bibinfo{volume}{30} (\bibinfo{year}{2017}).
\newblock


\bibitem[Wang et~al\mbox{.}(2017)]%
        {wang2017deep}
\bibfield{author}{\bibinfo{person}{Ruoxi Wang}, \bibinfo{person}{Bin Fu},
  \bibinfo{person}{Gang Fu}, {and} \bibinfo{person}{Mingliang Wang}.}
  \bibinfo{year}{2017}\natexlab{}.
\newblock \showarticletitle{Deep \& cross network for ad click predictions}.
\newblock In \bibinfo{booktitle}{\emph{Proceedings of the ADKDD'17}}.
  \bibinfo{pages}{1--7}.
\newblock


\bibitem[Wang et~al\mbox{.}(2021)]%
        {wang2021dcn}
\bibfield{author}{\bibinfo{person}{Ruoxi Wang}, \bibinfo{person}{Rakesh
  Shivanna}, \bibinfo{person}{Derek Cheng}, \bibinfo{person}{Sagar Jain},
  \bibinfo{person}{Dong Lin}, \bibinfo{person}{Lichan Hong}, {and}
  \bibinfo{person}{Ed Chi}.} \bibinfo{year}{2021}\natexlab{}.
\newblock \showarticletitle{DCN v2: Improved deep \& cross network and
  practical lessons for web-scale learning to rank systems}. In
  \bibinfo{booktitle}{\emph{Proceedings of the Web Conference 2021}}.
  \bibinfo{pages}{1785--1797}.
\newblock


\bibitem[Xu et~al\mbox{.}(2022)]%
        {xu2022rethinking}
\bibfield{author}{\bibinfo{person}{Jiajing Xu}, \bibinfo{person}{Andrew Zhai},
  {and} \bibinfo{person}{Charles Rosenberg}.} \bibinfo{year}{2022}\natexlab{}.
\newblock \showarticletitle{Rethinking personalized ranking at {Pinterest}: An
  end-to-end approach}. In \bibinfo{booktitle}{\emph{Proceedings of the 16th
  ACM Conference on Recommender Systems}}. \bibinfo{pages}{502--505}.
\newblock


\bibitem[Yang et~al\mbox{.}(2020)]%
        {yang2020mixed}
\bibfield{author}{\bibinfo{person}{Ji Yang}, \bibinfo{person}{Xinyang Yi},
  \bibinfo{person}{Derek Zhiyuan~Cheng}, \bibinfo{person}{Lichan Hong},
  \bibinfo{person}{Yang Li}, \bibinfo{person}{Simon Xiaoming~Wang},
  \bibinfo{person}{Taibai Xu}, {and} \bibinfo{person}{Ed~H Chi}.}
  \bibinfo{year}{2020}\natexlab{}.
\newblock \showarticletitle{Mixed negative sampling for learning two-tower
  neural networks in recommendations}. In \bibinfo{booktitle}{\emph{Companion
  Proceedings of the Web Conference 2020}}. \bibinfo{pages}{441--447}.
\newblock


\bibitem[Yi et~al\mbox{.}(2019)]%
        {yi2019sampling}
\bibfield{author}{\bibinfo{person}{Xinyang Yi}, \bibinfo{person}{Ji Yang},
  \bibinfo{person}{Lichan Hong}, \bibinfo{person}{Derek~Zhiyuan Cheng},
  \bibinfo{person}{Lukasz Heldt}, \bibinfo{person}{Aditee Kumthekar},
  \bibinfo{person}{Zhe Zhao}, \bibinfo{person}{Li Wei}, {and}
  \bibinfo{person}{Ed Chi}.} \bibinfo{year}{2019}\natexlab{}.
\newblock \showarticletitle{Sampling-bias-corrected neural modeling for large
  corpus item recommendations}. In \bibinfo{booktitle}{\emph{Proceedings of the
  13th ACM Conference on Recommender Systems}}. \bibinfo{pages}{269--277}.
\newblock


\bibitem[Yu et~al\mbox{.}(2021)]%
        {supersonalization}
\bibfield{author}{\bibinfo{person}{Lucia Yu}, \bibinfo{person}{Ethan Benjamin},
  \bibinfo{person}{Congzhe Su}, \bibinfo{person}{Yinlin Fu},
  \bibinfo{person}{Jon Eskreis-Winkler}, \bibinfo{person}{Xiaoting Zhao}, {and}
  \bibinfo{person}{Diane Hu}.} \bibinfo{year}{2021}\natexlab{}.
\newblock \showarticletitle{Personalization in e-commerce product search by
  user-centric ranking}.
\newblock \bibinfo{journal}{\emph{KMECommerce Workshop Held at WWW ’21}}
  (\bibinfo{year}{2021}).
\newblock


\bibitem[Zhai and Wu(2018)]%
        {pinterest18classification}
\bibfield{author}{\bibinfo{person}{Andrew Zhai} {and} \bibinfo{person}{Hao{-}Yu
  Wu}.} \bibinfo{year}{2018}\natexlab{}.
\newblock \showarticletitle{Making classification competitive for deep metric
  learning}.
\newblock \bibinfo{journal}{\emph{CoRR}}  \bibinfo{volume}{abs/1811.12649}
  (\bibinfo{year}{2018}).
\newblock
\showeprint[arXiv]{1811.12649}
\urldef\tempurl%
\url{http://arxiv.org/abs/1811.12649}
\showURL{%
\tempurl}


\bibitem[Zhai et~al\mbox{.}(2019)]%
        {pinterest19unified}
\bibfield{author}{\bibinfo{person}{Andrew Zhai}, \bibinfo{person}{Hao{-}Yu Wu},
  \bibinfo{person}{Eric Tzeng}, \bibinfo{person}{Dong~Huk Park}, {and}
  \bibinfo{person}{Charles Rosenberg}.} \bibinfo{year}{2019}\natexlab{}.
\newblock \showarticletitle{Learning a unified embedding for visual search at
  {P}interest}.
\newblock \bibinfo{journal}{\emph{CoRR}}  \bibinfo{volume}{abs/1908.01707}
  (\bibinfo{year}{2019}).
\newblock
\showeprint[arXiv]{1908.01707}
\urldef\tempurl%
\url{http://arxiv.org/abs/1908.01707}
\showURL{%
\tempurl}


\bibitem[Zhai et~al\mbox{.}(2016)]%
        {zhai2016deepintent}
\bibfield{author}{\bibinfo{person}{Shuangfei Zhai}, \bibinfo{person}{Keng-hao
  Chang}, \bibinfo{person}{Ruofei Zhang}, {and} \bibinfo{person}{Zhongfei~Mark
  Zhang}.} \bibinfo{year}{2016}\natexlab{}.
\newblock \showarticletitle{Deepintent: Learning attentions for online
  advertising with recurrent neural networks}. In
  \bibinfo{booktitle}{\emph{Proceedings of the 22nd ACM SIGKDD international
  conference on knowledge discovery and data mining}}.
  \bibinfo{pages}{1295--1304}.
\newblock


\bibitem[Zhang et~al\mbox{.}(2022)]%
        {zhang2022dhen}
\bibfield{author}{\bibinfo{person}{Buyun Zhang}, \bibinfo{person}{Liang Luo},
  \bibinfo{person}{Xi Liu}, \bibinfo{person}{Jay Li}, \bibinfo{person}{Zeliang
  Chen}, \bibinfo{person}{Weilin Zhang}, \bibinfo{person}{Xiaohan Wei},
  \bibinfo{person}{Yuchen Hao}, \bibinfo{person}{Michael Tsang},
  \bibinfo{person}{Wenjun Wang}, {et~al\mbox{.}}}
  \bibinfo{year}{2022}\natexlab{}.
\newblock \showarticletitle{{DHEN}: A deep and hierarchical ensemble network
  for large-scale click-through rate prediction}.
\newblock \bibinfo{journal}{\emph{arXiv preprint arXiv:2203.11014}}
  (\bibinfo{year}{2022}).
\newblock


\bibitem[Zhang et~al\mbox{.}(2020)]%
        {zhang2020personalized}
\bibfield{author}{\bibinfo{person}{Mengqi Zhang}, \bibinfo{person}{Shu Wu},
  \bibinfo{person}{Meng Gao}, \bibinfo{person}{Xin Jiang}, \bibinfo{person}{Ke
  Xu}, {and} \bibinfo{person}{Liang Wang}.} \bibinfo{year}{2020}\natexlab{}.
\newblock \showarticletitle{Personalized graph neural networks with attention
  mechanism for session-aware recommendation}.
\newblock \bibinfo{journal}{\emph{IEEE Transactions on Knowledge and Data
  Engineering}} (\bibinfo{year}{2020}).
\newblock


\bibitem[Zhang et~al\mbox{.}(2021a)]%
        {zhang2021deep}
\bibfield{author}{\bibinfo{person}{Weinan Zhang}, \bibinfo{person}{Jiarui Qin},
  \bibinfo{person}{Wei Guo}, \bibinfo{person}{Ruiming Tang}, {and}
  \bibinfo{person}{Xiuqiang He}.} \bibinfo{year}{2021}\natexlab{a}.
\newblock \showarticletitle{Deep learning for click-through rate estimation}.
\newblock \bibinfo{journal}{\emph{arXiv preprint arXiv:2104.10584}}
  (\bibinfo{year}{2021}).
\newblock


\bibitem[Zhang et~al\mbox{.}(2021b)]%
        {zhang2021user}
\bibfield{author}{\bibinfo{person}{Yang Zhang}, \bibinfo{person}{Dong Wang},
  \bibinfo{person}{Qiang Li}, \bibinfo{person}{Yue Shen}, \bibinfo{person}{Ziqi
  Liu}, \bibinfo{person}{Xiaodong Zeng}, \bibinfo{person}{Zhiqiang Zhang},
  \bibinfo{person}{Jinjie Gu}, {and} \bibinfo{person}{Derek~F Wong}.}
  \bibinfo{year}{2021}\natexlab{b}.
\newblock \showarticletitle{User retention: A causal approach with triple task
  modeling.}. In \bibinfo{booktitle}{\emph{IJCAI}}.
  \bibinfo{pages}{3399--3405}.
\newblock


\bibitem[Zhao et~al\mbox{.}(2018)]%
        {etsy2018iteminteraction}
\bibfield{author}{\bibinfo{person}{Xiaoting Zhao}, \bibinfo{person}{Raphael
  Louca}, \bibinfo{person}{Diane Hu}, {and} \bibinfo{person}{Liangjie Hong}.}
  \bibinfo{year}{2018}\natexlab{}.
\newblock \showarticletitle{Learning item-interaction embeddings for user
  recommendations}.
\newblock \bibinfo{journal}{\emph{arXiv preprint arXiv:1812.04407}}
  (\bibinfo{year}{2018}).
\newblock


\bibitem[Zhao et~al\mbox{.}(2019)]%
        {zhao2019recommending}
\bibfield{author}{\bibinfo{person}{Zhe Zhao}, \bibinfo{person}{Lichan Hong},
  \bibinfo{person}{Li Wei}, \bibinfo{person}{Jilin Chen},
  \bibinfo{person}{Aniruddh Nath}, \bibinfo{person}{Shawn Andrews},
  \bibinfo{person}{Aditee Kumthekar}, \bibinfo{person}{Maheswaran
  Sathiamoorthy}, \bibinfo{person}{Xinyang Yi}, {and} \bibinfo{person}{Ed
  Chi}.} \bibinfo{year}{2019}\natexlab{}.
\newblock \showarticletitle{Recommending what video to watch next: a multitask
  ranking system}. In \bibinfo{booktitle}{\emph{Proceedings of the 13th ACM
  Conference on Recommender Systems}}. \bibinfo{pages}{43--51}.
\newblock


\bibitem[Zhou et~al\mbox{.}(2019)]%
        {zhou2019deep}
\bibfield{author}{\bibinfo{person}{Guorui Zhou}, \bibinfo{person}{Na Mou},
  \bibinfo{person}{Ying Fan}, \bibinfo{person}{Qi Pi}, \bibinfo{person}{Weijie
  Bian}, \bibinfo{person}{Chang Zhou}, \bibinfo{person}{Xiaoqiang Zhu}, {and}
  \bibinfo{person}{Kun Gai}.} \bibinfo{year}{2019}\natexlab{}.
\newblock \showarticletitle{Deep interest evolution network for click-through
  rate prediction}. In \bibinfo{booktitle}{\emph{Proceedings of the AAAI
  conference on artificial intelligence}}, Vol.~\bibinfo{volume}{33}.
  \bibinfo{pages}{5941--5948}.
\newblock


\bibitem[Zhou et~al\mbox{.}(2018)]%
        {zhou2018deep}
\bibfield{author}{\bibinfo{person}{Guorui Zhou}, \bibinfo{person}{Xiaoqiang
  Zhu}, \bibinfo{person}{Chenru Song}, \bibinfo{person}{Ying Fan},
  \bibinfo{person}{Han Zhu}, \bibinfo{person}{Xiao Ma},
  \bibinfo{person}{Yanghui Yan}, \bibinfo{person}{Junqi Jin},
  \bibinfo{person}{Han Li}, {and} \bibinfo{person}{Kun Gai}.}
  \bibinfo{year}{2018}\natexlab{}.
\newblock \showarticletitle{Deep interest network for click-through rate
  prediction}. In \bibinfo{booktitle}{\emph{Proceedings of the 24th ACM SIGKDD
  international conference on knowledge discovery \& data mining}}.
  \bibinfo{pages}{1059--1068}.
\newblock


\end{thebibliography}

\pagebreak

\appendix
\section{Additional Information for Reproducibility}

\subsection{Candidate Generation for Personalized Ad Ranking: Ads Retrieval}
\label{appendix:candidates}

 Since performing real-time inference for adSformer CTR and PCCVR models for the full set of ad listings would be prohibitively expensive, we briefly describe the candidate retrieval system that produce up to 600 candidate listings to re-rank. We employ a hybrid lexical and pretrained representation-based retrieval system, designed to produce maximally relevant results for all types of user queries. 
 \noindent\textbf{Lexical retrieval.} In this first retrieval pass, we use "static" CTR and PCCVR models that are batch inferenced offline daily. The predicted scores, along with every listing's title and tags, as well as the ad budget remaining for each listing's campaign, are indexed in a sharded inverted index search database running Apache Solr\footnote{https://solr.apache.org/}. At query time, we fetch up to 1000 listings from each shard based on title and tag matching and that have ad budget remaining, and we calculate a ranking score which is similar to the value function. We also boost listings by a taxonomy matching prediction score, obtained from a separately trained and batch inferenced BERT \cite{devlin2018bert} model, as well as additional business logic.
\noindent\textbf{Pretrained representation-based retrieval.} We currently leverage representations from a two-tower(query and listing towers, textual inputs only) model built for organic search retrieval at Etsy, similarly to Amazon's \cite{amazonNIR}. We have plans to test multimodal AIR representations next. Given the trained two-tower model, we host the query tower online for live inference and batch inference the representations from the listing tower, which are indexed into an Approximate Nearest Neighbor (ANN) database. A large challenge we faced adopting off-the-shelf ANN solutions for an ads use case is that the index quickly becomes stale as sellers' budgets deplete throughout the day, ANN options on the market do not allow for filtering on attributes that change with such high frequency. We are experimenting with promising new in-house infrastructure that incorporates integrated filtering directly into the ANN.

\subsection{Training the adSformer CTR and PCCVR: Additional Details}
\label{appendix:training}

We run extensive offline experiments and ablation studies and omit many of the results for brevity. For example, we study the vocabulary size effect when encoding the ADPM representations of listing ID sequences and give results in \ref{tab:metrics1}. All three components of the ADPM-learned, pretrained and adSformer encoded representations-use the same vocabulary. We gain minimally by increasing vocabulary size after a certain point.

\begin{table*}
  \caption{Performance metrics for two selected adSformer CTR models with vocabulary sizes of 600K and 750K vs  baseline NN, corresponding to training dynamics in Figures \ref{fig:comet1} and \ref{fig:comet2} from Comet}
  \label{tab:metrics1}
  \begin{tabular}{lrrrr}
    \toprule
    Model & PR AUC (lift range) & ROC AUC (lift range) & Train time (machine)\\
    \midrule
    adSformer CTR 750 (fuchsia) & \textbf{0.0425 (+6.15\% to +6.65)}&\textbf{0.73954 (+1.94 to +2.36\%)} & 11 hrs (A100)\\
    adSformer CTR 600 (purple)  & 0.0423 (+5.00\% to +5.75\%) &0.7391 (+1.83\% to +1.9\%) & 10 hrs (A100)\\
    baseline NN CTR (orange) & 0.0400  & 0.7258 & 16 hrs (P100)\\
  \bottomrule
\end{tabular}
\end{table*}

Note that throughout this article the offline comparison numbers may vary for the same model variant due to variability inherent in sampled datasets. For reproducibility, Table \ref{tab:configs1} gives all hyperparameters and other machine learning choices for the winning adSformer CTR.

\begin{table}
  \caption{adSformer CTR: 600K vs 750K vocabulary size}
  \label{tab:configs1}
  \tiny
  \begin{tabular}{lrrrr}
    \toprule
    Config & adSformer750 & adSformer600 & baseline NN\\
    \midrule  
    Num Epochs & 1 & 1 & 3 \\
    Learning Rate & 0.002 & 0.001 & 0.00061\\
    Learning Rate Scheduler & cosine & cosine & cosine\\
    Optimizer & Adam & Adam & LazyAdam\\
    Adam $\beta_1$ & 0.9 & 0.9 & 0.9 \\
    Adam $\beta_2$ & 0.999 & 0.999 & 0.999 \\
    Adam $\epsilon$ & 1e-08 & 1e-08 & 1e-08\\
    Dropout Hidden Layers & 0 & 0 & 0\\
    Batch Norm Hidden Layer& yes & yes & yes \\
    Cross Module (DCN) & yes & yes & no \\
    Batch Size (train) & 8192 & 8192 & 8192\\
    Batch Size (validation) & 500 & 500 & 500\\
    Machine Type (training) & A100 & A100 & P100\\
    Machine Type (evaluation) & A100 & P100 & T4\\
    Loss & BinaryCrossEntropy & BinaryCrossEntropy & BinaryCrossEntropy\\
    Num Parameters (trainable) & 708,504,031 & 690,842,559 & 233,709,623\\
    Num Parameters (total) & 897,499,741 & 844,564,093 & 243,726,397 \\
  \bottomrule
    \end{tabular}
\end{table}

Table \ref{tab:configs_data} gives the dataset choices for the winning adSformer CTR and PCCVR. For the CTR we sample 50 percent of non-clicked impressions in the training dataset but perform a random sampling for the validation set, so that validation and test sets are representative of the production data seen in the wild. The PCCVR is trained on clicked impressions but evaluated on a representative test sample, similarly to the adSformer CTR.

\begin{table}
  \caption{Train and validation datasets characteristics for adSformer CTR and PCCVR}
  \label{tab:configs_data}
  \footnotesize
  \begin{tabular}{lrrrr}
    \toprule
    Model & Config & Training & Validation \\
    \midrule  
    CTR & Num Examples & ~300mln & ~8.5mln \\
    CTR & Num Consecutive Days & 30 & 1 \\
    CTR & Sampling Type & 50-50 negative sampling & random sampling \\
    PCCVR & Num Examples & ~200mln & ~8.5mln \\
    PCCVR & Num Consecutive Days & 21 & 1\\
    PCCVR & Sampling Type & click filtered & random sampling \\
  \bottomrule
    \end{tabular}
\end{table}

Furthermore, Table \ref{tab:pccvr_train_window} illustrates gains in PCCVR performance from increased training dataset size after including the ADPM.

\begin{table}[h]
  \caption{Increased performance metrics for adSformer PCCVR model vs the baseline NN production model on the same validation dataset for different training window sizes}
  \label{tab:pccvr_train_window}
  \small
  \begin{tabular}{lrrrr}
    \toprule
    Training Window & ROC-AUC Lift & PR-AUC Lift & Train Duration \\
    \midrule
    2 weeks Baseline & 0.0\% & 0.0\% & 6 hrs\\
    3 weeks Baseline & +0.01\% & +0.11\% & 8 hrs\\
    \hline
    2 weeks adSformer & +1.28\% & +9.61\% & 11 hrs \\
    3 weeks adSformer & +1.58\% & +12.07\% & 16 hrs \\
  \bottomrule
\end{tabular}
\end{table}

\begin{figure}[h]
  \centering
  \includegraphics[width=\linewidth]{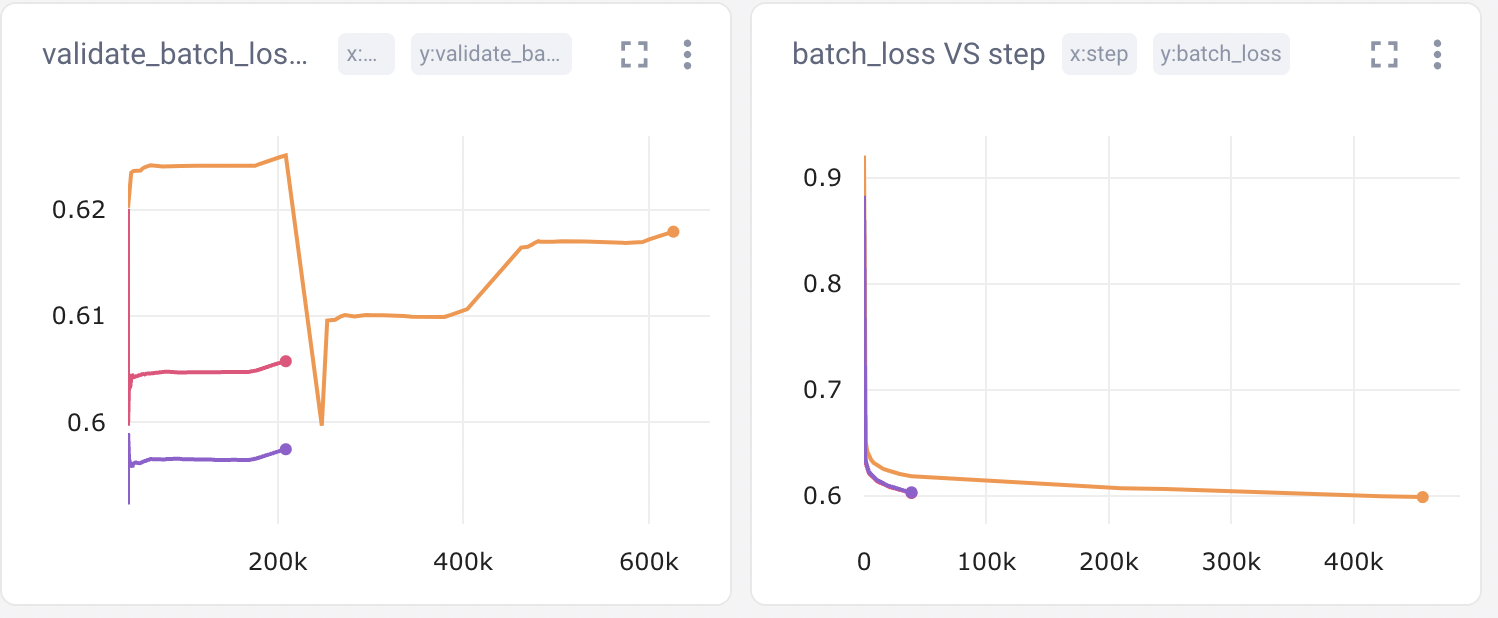}
  \caption{adSformer CTR model selection: train and validation loss curves. Lower is better. Source: \href{https://www.comet.com/site/}{Comet.}}
  \label{fig:comet1}
\end{figure}

\begin{figure}[h]
  \centering
  \includegraphics[width=\linewidth]{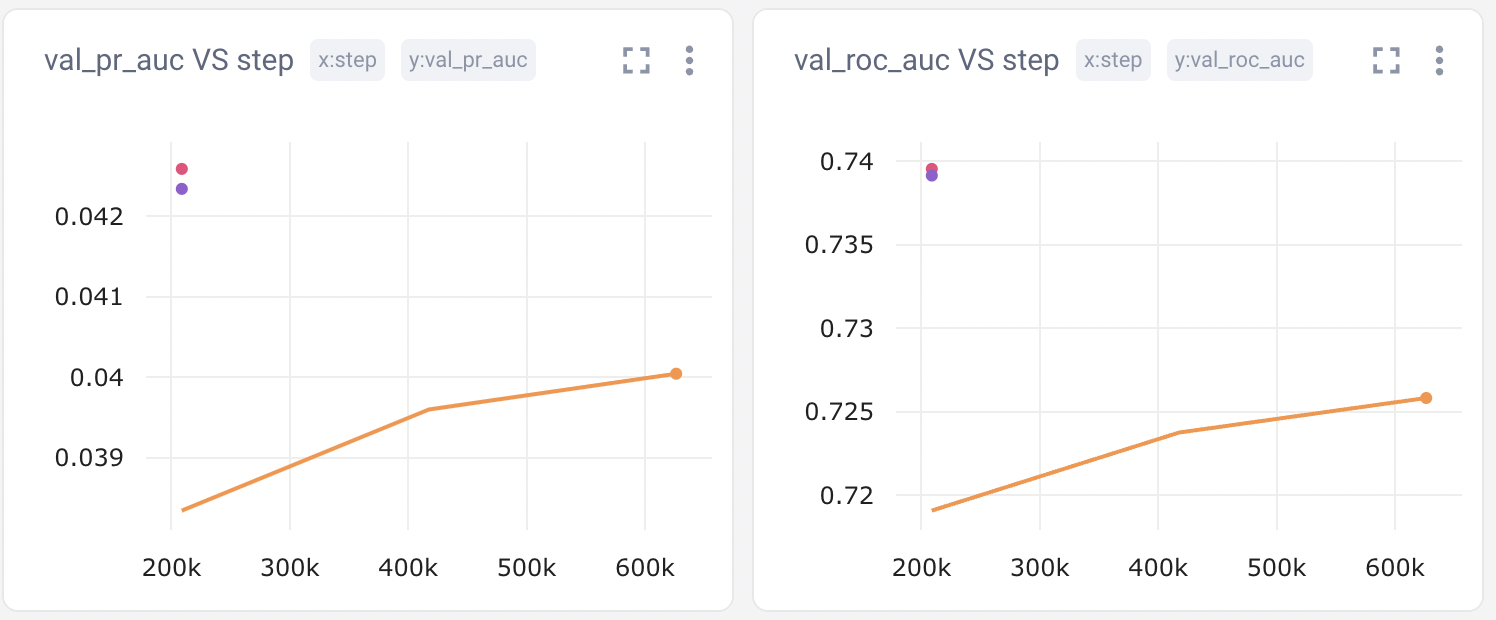}
  \caption{adSformer CTR selected models vs baseline: ROC AUC and PR AUC curves. Higher is better. Source: \href{https://www.comet.com/site/}{Comet.}}
  \label{fig:comet2}
\end{figure}

To determine optimal hyperparameters, we combine random search, deep learning expert manual selection, and Bayesian hyperparameter search, depending on the hyperparameter of interest. We visualize training dynamics curves in Comet \cite{CometML} to guide our model choices. For example, in Figures \ref{fig:comet1} we see that both adSformer CTR model flavors (with listing ID vocabularies of 750K and 600K respectively) trained for one epoch achieve slightly lower training losses and significantly lower validation losses when compared to the baseline neural network trained for three epochs. Table \ref{tab:configs1} gives the complete configuration for these three compared models. In Table \ref{tab:metrics1} we see the final metric lifts and training times achieved on the same train and validation datasets (described in Table \ref{tab:configs_data}). Figure \ref{fig:comet2} helps visually understand the lift achieved by the adSformer CTR variants in validation AUC metrics.

\begin{figure}[h]
  \centering
  \includegraphics[width=\linewidth]{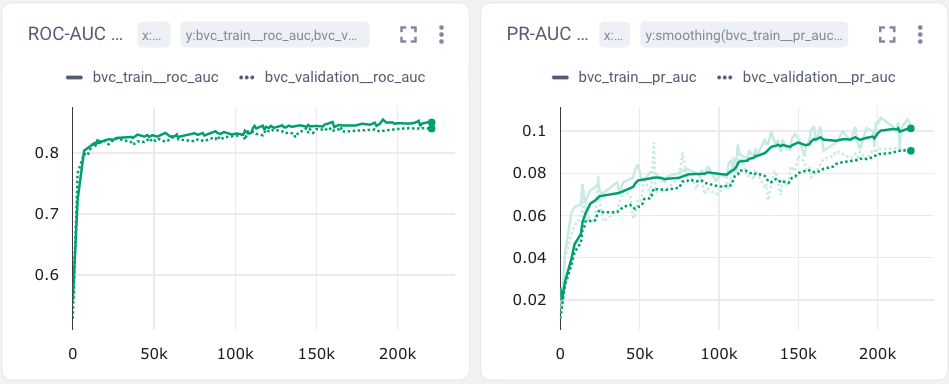}
  \caption{adSformer PCCVR eval metrics during training: ROC AUC and PR AUC per step. Higher is better. Source: \href{https://www.comet.com/site/}{Comet.}}
  \label{fig:comet3}
\end{figure}

In additional ablation studies  we vary multiple hyperparameters, for example the number of heads in the adSformer encoder and the listing representation size in ADPM's learned component. In Table \ref{tab:additional} we see that removing the average pooled learned listing representation from ADPM in adSformer CTR leads to a drop in ROC AUC but an increase in PR AUC. We have a better idea of online to offline correlations between offline test ROC AUC and online CTR, so we select the increased ROC AUC model. Lowering the size of the learned listing representations has a similar effect. The addition of one head to the adSformer encoder component of ADPM increases both metrics and so does the removal of dropout. 

\begin{table}
  \caption{Additional ablations studies. Offline metric lift in comparison to the baseline for the adSformer CTR.}
  \label{tab:additional}
  \footnotesize
  \begin{tabular}{lrrrr}
    \toprule
    Config & Test ROC AUC Lift & Test PR AUC Lift \\
    \midrule  
    All Listing Pretrained + Listing Learned & \textbf{+2.26\%} &  +6.17\%\\
    All Listing Pretrained + No Listing Learned & +2.18\% &  \textbf{+6.20\%}\\
    \midrule
    adSformer Encoder 2 Heads & +2.13\% &  +6.42\%\\
    adSformer Encoder 3 Heads & \textbf{+2.14\%} &  \textbf{+6.44\%}\\
    \midrule
    Learned Listing Size 16 &+2.10\% &  +6.50\%\\
    Learned Listing Size 32 &+2.14\% &  +6.44\%\\
    Learned Listing Size 64 &+2.04\% &  +6.14\%\\
    \midrule
     adSformer Encoder 0 Dropout & \textbf{+2.14\%} &  \textbf{+6.44\%}\\
     adSformer Encoder 0.1 Dropout & \textbf{+2.10\%} &  \textbf{+6.34\%}\\
  \bottomrule
    \end{tabular}
\end{table}

We performed similar ablation studies for the adSformer PCCVR model and Table \ref{tab:configs2} depicts the final hyperparameter settings for the winning adSformer PCCVR.

\begin{table}
  \caption{adSformer PCCVR hyperparameters and other algorithmic choices}
  \label{tab:configs2}
  \footnotesize
  \begin{tabular}{lrrrr}
    \toprule
    Config & adSformer PCCVR & baseline NN\\
    \midrule  
    Num Epochs & 2 & 2 \\
    Learning Rate & 0.0001 & 0.0001\\
    Learning Rate Scheduler & cosine & cosine\\
    Optimizer & Adam  & LazyAdam\\
    Adam $\beta_1$ & 0.9 & 0.9 \\
    Adam $\beta_2$ & 0.999 & 0.999\\
    Adam $\epsilon$ & 1e-08 & 1e-08\\
    Dropout Hidden Layers & 0.02 & 0\\
    Batch Norm Hidden Layer& yes & yes\\
    Cross Module (DCN) & yes & yes \\
    Batch Size (train) & 500 & 500\\
    Batch Size (validation) & 500 & 500\\
    Machine Type (training) & P100 & P100\\
    Machine Type (evaluation) & P100 & T4\\
    Loss & BinaryCrossEntropy & BinaryCrossEntropy\\
    Num Parameters (trainable) & 549,005,007 & 322,975,106\\
    Num Parameters (total) & 704,469,589 & 332,978,320\\
    Training Dataset Size & ~200M & ~135M\\
  \bottomrule
    \end{tabular}
\end{table}

Finally, Table \ref{table:final} presents one more set of the winning adSformer CTR and PCCVR offline performance results against production baseline NN.

\begin{table}
  \caption{Performance metrics for the selected adSformer CTR and PCCVR models vs NN baselines. Results extracted from the daily training and evaluation jobs.}
  \label{tab:offline_metrics}
  \small
  \begin{tabular}{lrrrr}
    \toprule
    Model & PR AUC & ROC AUC & Train time\\
    \midrule
    Baseline CTR NN  & 0.0411 & 0.719 & 14 hrs (1 P100))\\
    Baseline PCCVR NN  & 0.0602 & 0.824 & 8 hrs (1 P100))\\
    \midrule
    adSformer CTR & \textbf{+6.65\%} & \textbf{+2.16\%} & 11 hrs (1 A100) \\
    adSformer PCCVR & \textbf{+12.70\%} & \textbf{+1.81\%} & 16hrs (1 P100)\\
  \bottomrule
  \label{table:final}
\end{tabular}
\end{table}

\subsection{Additional Sampling Biases Discussion}
\label{sec:bias-appendix}
 In addition to position debiasing, we are actively considering the practical implications of other selection biases, such as the conditionality of our PCCVR prediction target on the CTR target. The personalized PCCVR model predicts $P(y_{PCCVR}=1|y_{CTR}=1, x)$ effectively ignoring the sample space of non-clicked purchased items in the personalized search ads space, the case of $P(y_{PCCVR}=1|y_{CTR}=0, x)$. In preliminary evaluations of potential explicit treatment for this selection bias, for example by using an inverse propensity weighting in a multitask CTR-PCCVR formulation similarly to \cite{zhang2021user}, or the approach in \cite{ma2018entire}, we determined that the metric improvements do not sufficiently reward an increase in machine learning system complexity. We plan to revisit sampling biases in the future, possibly as part of a multitask learning paradigm.

\subsection{Additional Calibration Information}

After the uncalibrated model is trained, we use the raw logits as a feature to the calibration layer, which is a simple logistic regression model trained on the validation set reflecting the production data. The calibration layer learns a single parameter $A$ and bias term $B$, and outputs $P\left(y=1|x\right)={\frac {1}{1+\exp(-(Af(x)+B))}}$. The calibration mapping function is isotonic (monotonically increasing), preserving the relative ordering of predictions. As a sanity-check, the evaluation metric, AUC, should be identical between the calibrated and uncalibrated models. We evaluate miscalibration by monitoring the Expected Calibration Error (ECE) and Normalized Cross Entropy (NCE). ECE aims to measure the difference in expectation between confidence and accuracy. We approximate the ECE by partitioning predictions into $M$ equally-spaced bins and taking the weighted average of the bins' $accuracy/confidence$ difference.

\subsection{Visual Representations Additional Information}

\subsubsection{Multitask Learning of Visual Representations}
We saw best results by training a model on the following four datasets/classification tasks: 1. listing images to top level taxonomy. 2. listing images to fine grained taxonomy. 3. listing images to seller-input primary color. 4. user-uploaded review photos to fine grained taxonomy. For the top level taxonomy and primary color tasks we sampled 16K images for a set of 15 labels each. For fine grained taxonomy we sampled 200 images for a set of 1000 taxonomy nodes. We replaced the final convolutional layer of the EfficienetNetB0 backbone with a 256-dimension layer to output representations of the same size. The model was trained by first freezing the backbone and training only classification heads for one epoch using a 0.001 learning rate, and then unfreezing the final 50 layers of the backbone and training for an additional 8 epochs using the same learning rate. We used Adam optimizer with values of 0.9, 0.999 and 1.0e-7 for beta1, beta2 and epsilon respectively.

\begin{figure}[h]
\includegraphics[width=3.25in]{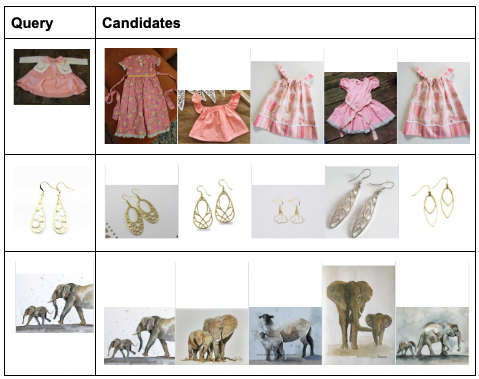}
\centering
\caption{Additional example of candidate retrieval using multitask visual representations. Left column: query image used. Right column: top-5 nearest neighbors retrieved}
\label{fig:viz-retrieval-examples}
\end{figure}

\subsubsection{ Ads Computer Vision: Visually Similar Ads Additional Information}

The listing-to-listing experience does not require real-time predictions, so we implemented a daily batch inference pipeline to generate image representations for our entire listing inventory. Our pipeline first extracts the primary image for each Etsy listing, then passes those forward through the frozen multitask visual representation model to generate representations. Representations for all listings are saved to a scalable key-value store. representations are also filtered down by active ad campaigns. These candidate representations are indexed into an \textit{Approximate Nearest Neighbor} (ANN) inverted file (IVF) index. IVF approximates a nearest neighbor search by first splitting the representation space to N clusters, and at inference time only searching the K nearest clusters. Thus we can efficiently search the large space of representations.

\subsubsection{ Ads Computer Vision: Search by Image}
\label{sec:search_by_image:appendix}
Another application of the multitask visual representations was a brand new \textit{search by image} shopping experience, built by the Etsy Ads team, and available on the BOE app. Users can search using photos taken with their phone. We leverage the multitask dataset sampler to train on an additional task to classify review photos to their corresponding listing’s taxonomy. Review photos are photos of purchased items taken “in the wild” by users. Our hypothesis is that these images should serve as a proxy for user-taken query images.

The search by image experience requires generating the representation of the query image on the fly. To enable fast inference on a large CNN such as the visual representations model, we partnered with the ML platform team to deploy the representation model as the first GPU service in production at Etsy.

Similarly to the visually similar ads module, the candidate representations of the search by image retrieval system are pre-computed by generating representations for primary listing images on a recurring basis. However, since we are serving ads as well as organic results, the ANN is indexed with the full inventory of nearly 100M active Etsy listings. At query time, the user-uploaded image is inferenced on the GPU-backed model to compute the representation, which is used to search the ANN index for the most visually similar listings.

\subsection{Multimodal Representations: Learning AIR Additional Information}
In addition to parameterizing the ADPM module in adSformer CTR use-case, we use AIR representations to retrieve recommendation-style ads for listing-to-listing requests, across various types of pages on both mobile and web. For example, the sash of sponsored recommendations appearing at the bottom of the listing page is backed by AIR representations. In listing-to-listing requests candidates are retrieved using a \textit{query} representation generated from the features of the viewed listing. Typically, we refer to the listing used to generate the query representation as the \textit{source} listing.

\subsubsection{Additional Training Details}
Our final representation model was trained using 30M of historical click pairs, from a look-back window of 45 days. For training we used the Adam optimizer, with a learning rate, beta1, beta2, and epsilon values of 0.0001, 0.9, 0.999, and 1e-7, respectively. We trained using a batch size of 4096, and sample 1500 negative labels to calculate the loss.

\subsubsection{Evaluating AIR: Online Results}
Online A/B experiments were run against the control of neural IR representations extracted from the listing tower of a model trained on search purchases (a less relevant objective for ads). Two treatments were tested to measure the effect of different dimensions of the input text representations, 128d and 256d, since the text representation lookup table makes for the majority of trainable parameters in the overall AIR model. Table \ref{table:AIR_online} show online CTR gains.

\begin{table}
    \centering
    \caption{Online AIR A/B experiment results showing lift in CTR.}
    \label{tab:air-exp}
    \begin{tabular}{crrr}
        \toprule
        fastText Dimension & CTR Lift\\
        \midrule
        128d & +2.79\%\\
        256d & +2.88\%\\        
        \bottomrule
    \end{tabular}
    \label{table:AIR_online}
\end{table}

\subsection{Text Representations Additional Information}
Raw text inputs are first preprocessed to standardize casing and punctuation, mask stop-words and numbers, and remove extraneous spaces and symbols. To reduce the size of the final lookup table we filter out tokens which have less than 10 occurrences. To maintain relevance to the ad use case, we keep tokens which appeared at least once in the last 30 days of ad interactions even if they occur less than 10 times in the data. The final lookup table contains roughly one million tokens. 

We trained the skip-gram word representations for five epochs. We used a learning rate of 0.05, a context window of five, and a minimum and maximum character n-grams lengths of three and six respectively. We trained using negative sampling with five sampled negatives. We experimented with outputting both 128 and 256 dimension representations. We have found that the larger representations perform slightly better in most downstream use cases. However, the number of trainable parameters of the lookup table grows significantly with larger representations, and thus requires more expensive infrastructure to fine tune during training of downstream task, which we have seen in the adSformer CTR case.

\subsection{Skip-Gram Representations Additional Information}
\label{sec:skipgram-appendix}

The objective of the model is to predict, given a listing $l_i$ in the corpus, the probability that another listing $l_{i+j}$ will be observed within a fixed-length contextual window. The probability $p(l_{i+j} | l_i)$ is defined mathematically according to the softmax formula

\begin{equation}
p(l_{i+j} | l_i) = \frac{\exp(\mathbf{v}_{l_i}^\mathsf{T}\mathbf{v}_{l_{i+j}}')}{\sum_{k=1}^{|\mathcal{V}|}\exp(\mathbf{v}_{l_i}^\mathsf{T}\mathbf{v}_{l_{k}}')},
\end{equation}

where $\mathbf{v_l}$ and $\mathbf{v_l'}$ are the input and output vector representations of a listing $l$, respectively, and $\mathcal{V}$ is the vocabulary of unique listings. 

In our training data, we use two months' worth of user sessions, and we upsample sessions that included a purchase at a rate of 5:1. We chose representation dimension $d=64$ after observing that higher dimensions tend to improve representation quality but with diminishing returns and with trade offs in model training and inference cost. We use a context window size of five. Interestingly, increasing the window size had a positive impact for negative sampling models but had mixed results for hierarchical softmax models.

\begin{figure}[h]
\includegraphics[width=3.25in]{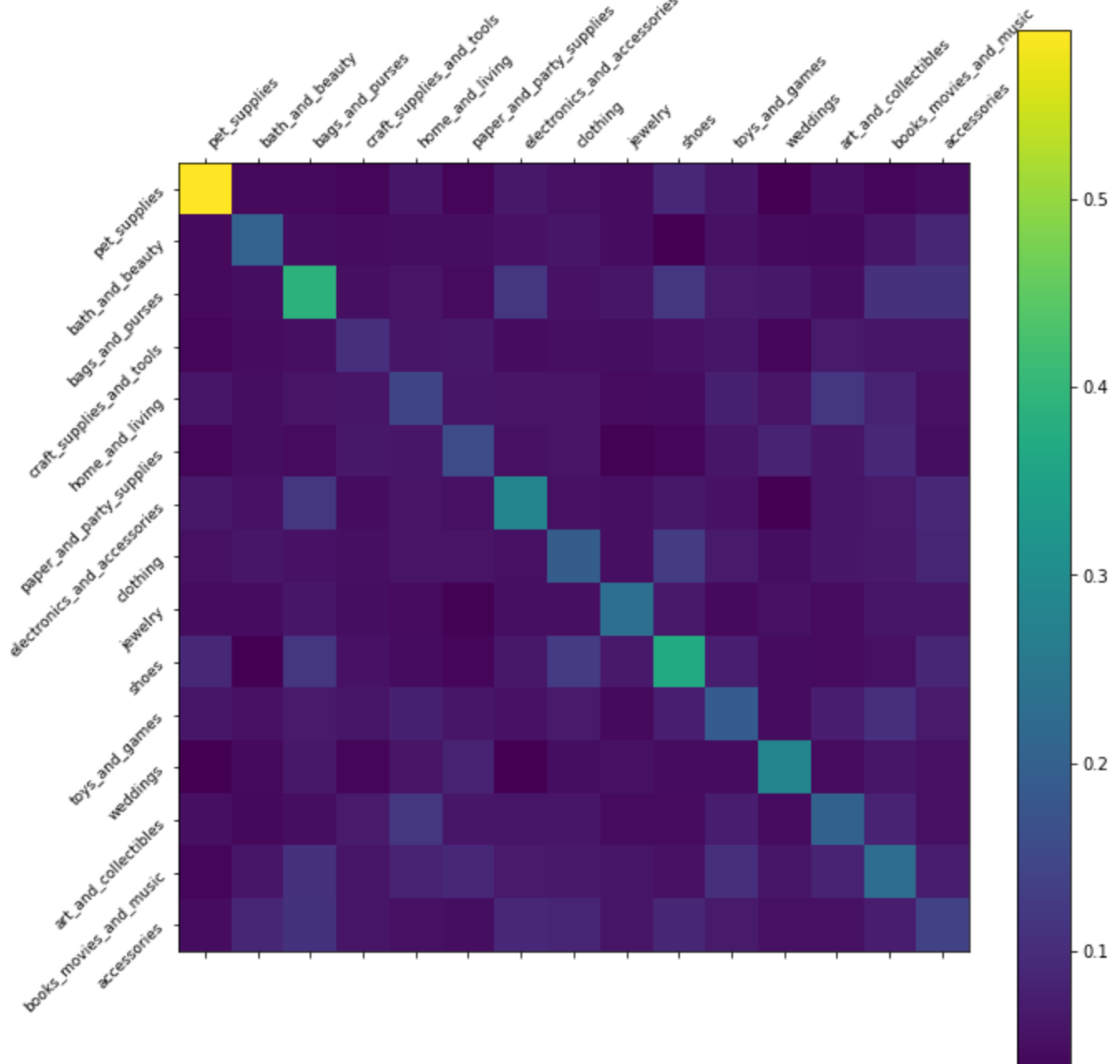}
\centering
\caption{The skip-gram listing representation model nearly perfectly encodes listing taxonomy.}
\label{fig:skip-gram-taxo}
\end{figure}

As a gut check on representation quality and to better understand what signals our representations were learning, we compared the cosine similarity of groups of listings segmented by various listing attributions. For example, listings which are co-viewed in a session are likely to belong to the same taxonomy node, so we would expect listings with the same taxonomy to have a higher cosine similarity than those with different taxonomies. Figure \ref{fig:skip-gram-taxo} shows that our skip-gram representations learned with hierarchical softmax nearly perfectly captured the listing taxonomy attribute.

To evaluate our representations, we ran batch data inference jobs with Apache Beam and Dataflow to compute the search ranking and attribute cosine similarity metrics, which we visualized in Colab notebooks.

\end{document}